\documentclass[11pt, a4paper]{article}

\usepackage[utf8]{inputenc}
\usepackage[T1]{fontenc}
\usepackage{amsmath}          
\usepackage{amssymb}          
\usepackage{graphicx}         
\usepackage[a4paper, margin=1in]{geometry} 
\usepackage{hyperref}         
\usepackage{booktabs}         
\usepackage{caption}          
\usepackage{float}            
\usepackage{amsthm}           

\newtheorem{definition}{Definition}
\usepackage{authblk}          
\usepackage{multirow}         
\usepackage{animate}          
\usepackage{xcolor}
\usepackage{bookmark}

\title{DeePoly: A High-Order Accuracy Scientific Machine Learning Framework for Function Approximation and Solving PDEs}
\author[1]{Li Liu}
\author[1]{Heng Yong}
\affil[1]{Institute of Applied Physics and Computational Mathematics, Beijing, China}

\renewcommand{\thanks}[1]{\footnotemark}
\date{\today\thanks{}}

\begin{document}

\maketitle
\footnotetext{Corresponding authors: \texttt{liu\_li@iapcm.ac.cn}, \texttt{yong\_heng@iapcm.ac.cn}}

\begin{abstract}
  Recently, machine learning methods have gained significant traction in scientific computing, particularly for solving Partial Differential Equations (PDEs). However, methods based on deep neural networks (DNNs) often lack convergence guarantees and computational efficiency compared to traditional numerical schemes. This work introduces DeePoly, a novel framework that transforms the solution paradigm from pure non-convex parameter optimization to a two-stage approach: first employing a DNN to capture complex global features, followed by linear space optimization with combined DNN-extracted features (Spotter) and polynomial basis functions (Sniper). This strategic combination leverages the complementary strengths of both methods---DNNs excel at approximating complex global features (i.e., high-gradient features) and stabilize the polynomial approximation while polynomial bases provide high-precision local corrections with convergence guarantees. Theoretical analysis and numerical experiments demonstrate that this approach significantly enhances both high-order accuracy and efficiency across diverse problem types while maintaining mesh-free and scheme-free properties.
 This paper also serves as a theoretical exposition for the open-source project \href{https://github.com/bfly123/DeePoly}{\textbf{DeePoly}} \includegraphics[width=2cm]{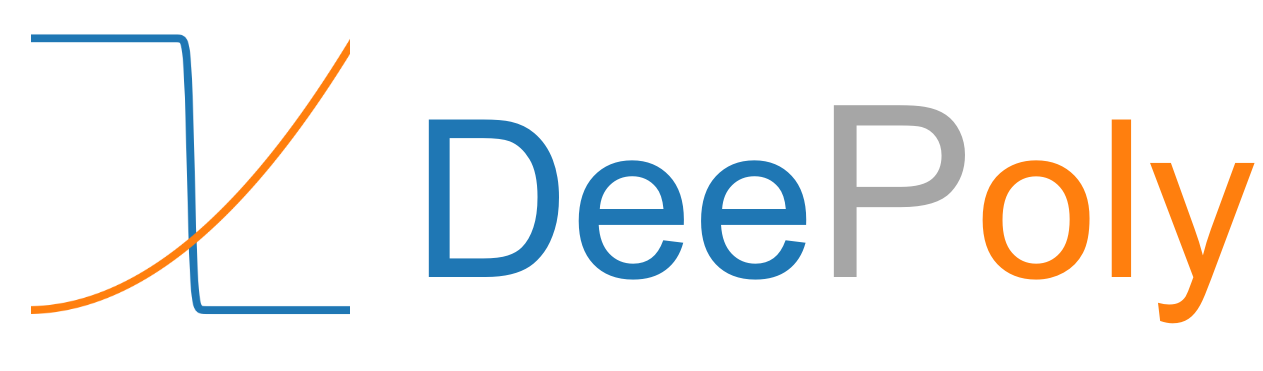}.
\end{abstract}

\vspace{0.5cm}
\begin{center}
    \textbf{Graphical Abstract}\\
    \vspace{0.3cm}
    \fbox{\includegraphics[width=0.98\textwidth]{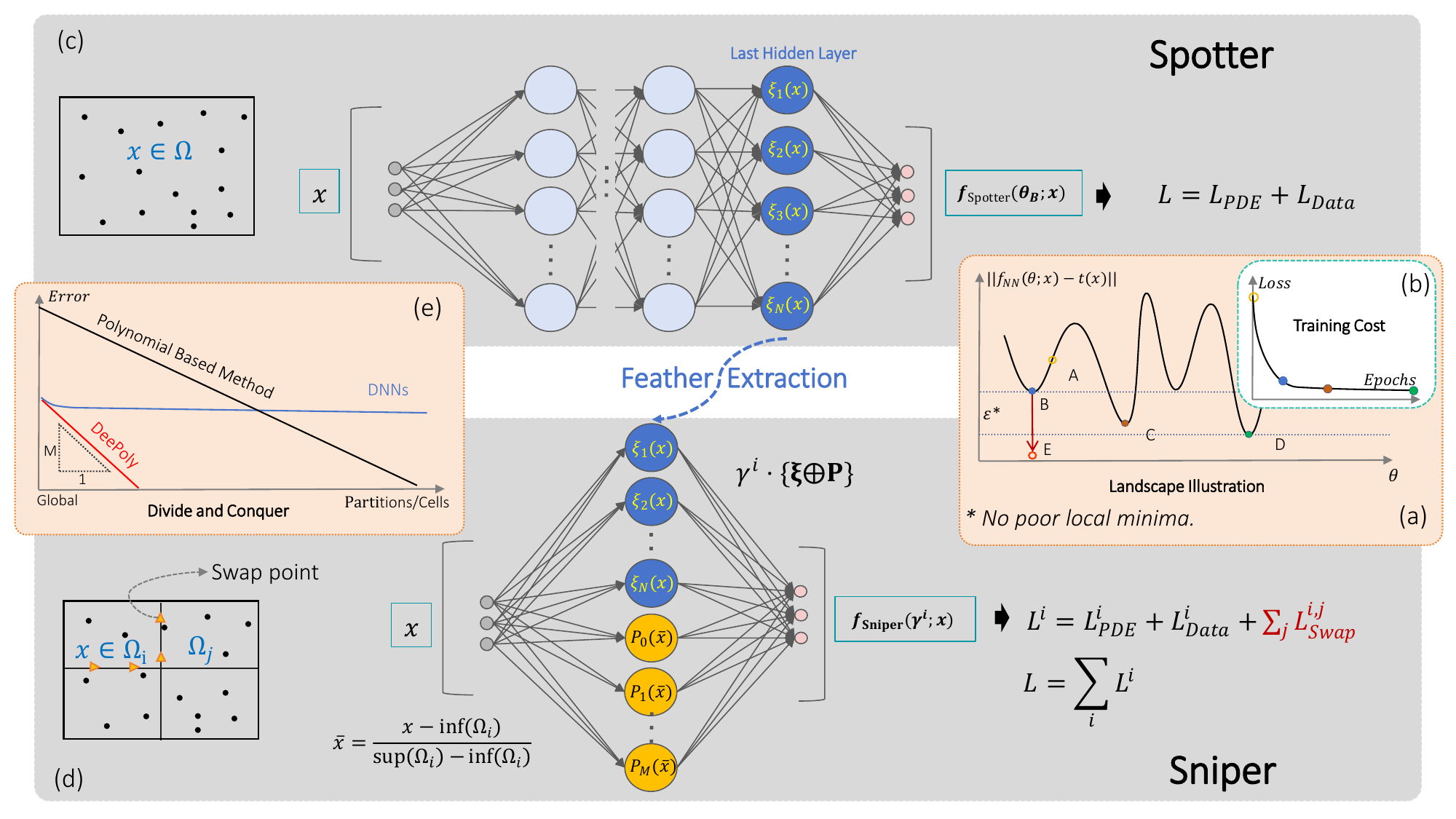}}
\end{center}
\vspace{0.5cm}


\section{Introduction}
Deep learning, empowered by the remarkable non-linear approximation capabilities of deep neural networks (DNNs) \cite{hornik1989multilayer,cybenko1989approximation}, has emerged as a transformative paradigm in scientific computing. Among various deep learning approaches, Physics-Informed Neural Networks (PINNs) \cite{raissi2019physics} have attracted substantial attention across engineering disciplines as an innovative methodology for solving Partial Differential Equations (PDEs). These methods offer promising alternatives to traditional numerical approaches such as the Finite Element Method (FEM) and Finite Difference Method (FDM) \cite{pinkus1999approximation}, particularly when addressing challenges that conventional methods struggle with. The widespread interest in PINNs stems from their distinctive advantages: they operate without requiring explicit mesh generation, eliminate the need for discretization schemes, and demonstrate remarkable versatility in handling both forward and inverse problems \cite{raissi2019physics}. Despite these compelling features, PINN-based approaches encounter significant challenges in achieving the error convergence rates and computational efficiency characteristic of established numerical methods \cite{karniadakis2021physics,wang2022when}. Quantitative studies have shown that even for relatively simple benchmark problems, PINNs typically require orders of magnitude more computational resources while achieving lower accuracy compared to traditional numerical solvers \cite{karniadakis2021physics,wang2022when}. These limitations, which arise from fundamental theoretical constraints in DNN approximation theory \cite{maiorov1999lower} and the inherent difficulties of non-convex optimization \cite{zhou2022towards}, have substantially restricted the broader adoption of PINNs in practical engineering applications.

Given that accuracy and efficiency constitute fundamental criteria for any mature numerical scheme, substantial research has focused on addressing these two critical limitations in neural network-based approaches. These efforts can be broadly categorized into methods targeting accuracy enhancement and those aiming to improve computational efficiency.

For accuracy enhancement, several strategies have emerged: (1) Advanced optimization techniques, including adaptive learning rate methods and second-order optimizers \cite{kingma2014adam,bottou2018optimization,wang2025gradient}, which can improve solution accuracy but often at significantly increased computational cost; (2) Modified network architectures \cite{wang2021understanding, jagtap2020adaptive, wang2022when} and high-frequency feature incorporation \cite{tancik2020fourier, wang2020eigenvector}, which can enhance fitting accuracy for solutions containing high-frequency structures, though typically only elevating challenging problems to accuracy levels comparable with low-frequency problems; and (3) Multi-stage neural network methods \cite{wang2024multi}, which iteratively approximate residuals and have demonstrated machine-precision accuracy but may face limitations in efficiency and generalization to complex non-linear systems.

For efficiency improvement, approaches include: (1) Shallow random networks such as Extreme Learning Machines (ELM) \cite{huang2006extreme, huang2015trends,dong2025hierachical} and Random Feature Models (RFM) \cite{chen2024optimization}, which can simultaneously improve accuracy and efficiency for smooth problems but often exhibit insufficient approximation capabilities for complex problems especially with large gradients or discontinuities; (2) Domain decomposition techniques \cite{li2020deep, jagtap2021extended}, which parallelize computation across subdomains but introduce challenges in maintaining solution continuity across boundaries; and (3) Alternative network formulations like Kolmogorov-Arnold Networks (KAN) \cite{liu2024kan}, which offer promising theoretical properties but remain in early stages and didn't show accuracy improvement than PINNs in PDE applications \cite{shukla2024comprehensive}.

Despite these extensive efforts, a fundamental challenge persists: no existing method consistently achieves both the guaranteed error convergence rates of traditional numerical schemes and comparable computational efficiency across a wide range of benchmark problems. Theoretical analysis demonstrates that this limitation stems from two fundamental issues: (1) the inherent difficulty of non-convex optimization in finding globally optimal solutions \cite{bottou2018optimization, sun2020optimization}, and (2) a previously under-explored limitation in DNNs' ability to accurately approximate even low-order polynomial functions with high precision. This latter deficiency is particularly consequential as it prevents DNNs from fully benefiting from the classical "divide and conquer" strategy that forms the theoretical foundation of traditional numerical methods' convergence properties. Consequently, while DNNs excel at approximating complex non-linear functions globally, they struggle to achieve the systematic error reduction and convergence guarantees that characterize mature numerical approaches.

To address these fundamental limitations, we propose a novel hybrid framework that strategically combines the complementary strengths of neural networks and classical approximation theory. Our approach introduces a two-stage solution strategy: first employing a DNN to capture complex global features of the solution, followed by a linear space optimization stage utilizing a combined space of DNN-extracted basis functions (Spotter) and polynomial basis functions (Sniper). The DNN component provides the approximation power necessary for complex solution features and stabilize the polynomial oscillation, while the polynomial basis enables the high-order precision local corrections required for guaranteed convergence properties.

{\color{red}
From an optimization perspective for DNNs, research inspired by the p-spin glass model in physics \cite{bray2007statistics,auffinger2013random} reveals that their loss landscapes are largely free of "poor" local minima \cite{choromanska2015loss,kawaguchi2016deep}. In other words, the performance gap between most local minima and the global optimum is small, which allows algorithms to efficiently find a good solution near the initial parameters.
However, this negligible difference from the perspective of network training may prove fatal in scientific computing---for instance, the distinction between \(10^{-3}\) and \(10^{-5}\) errors. Such fundamental divergence in precision tolerance across domains compels us to seek better local optima. The true challenge lies in the fact that while these high-quality local minima are virtually identical in performance, they may lie extremely far apart in parameter space. Consequently, any attempt to explore between these distant solutions to extract marginal performance gains inevitably leads to a dramatic surge in training costs.
Based on the above rationale, this paradigm fundamentally transforms the solution process: rather than relying solely on challenging non-convex parameter optimization, it adopts a mathematically tractable two-stage approach. First, the Spotter step efficiently identifies an acceptable solution near the initial parameters-achieving potentially substantial loss reduction (without necessarily converging to local minima) through computationally lightweight optimization. Second, the Sniper step employs high-order polynomial bases to refine accuracy via linear subspace optimization.
Theoretical analysis and numerical experiments demonstrate that this framework not only significantly enhances both accuracy and efficiency compared to pure DNN methods but also inherits the rigorous convergence guarantees of polynomial approximation while preserving key advantages of Physics-Informed Neural Networks (PINNs), including broad applicability, stability, mesh-free operation, and scheme independence. }

The remainder of this paper is organized as follows: Section 2 introduces the algorithmic approach, Section 3 presents comprehensive numerical examples across various problem types, Section 4 provides theoretical analysis of the method's convergence properties, and Section 5 concludes with a summary of contributions and directions for future research. The accompanying implementation, \href{https://github.com/bfly123/DeePoly}{DeePoly}, is made available as an open-source project to facilitate reproducibility and further development by the scientific computing community.

\section{Methodology}
\begin{quote}
    ``Sniper + Spotter > Sniper or Spotter''
\end{quote}
In military tactics, a sniper team typically consists of at least two members: a sniper and a spotter. The spotter is responsible for monitoring the tactical environment, identifying and prioritizing targets, and providing critical data such as distance and wind conditions. The sniper focuses on executing precise shots and maintaining stealth. Their complementary roles create a synergy that exceeds their individual capabilities.

This military analogy effectively illustrates our approach: Deep Neural Networks (DNNs) excel at capturing complex global features but struggle with local high-precision characterization. Conversely, continuous polynomial functions excel at local high-precision characterization but are less adept at capturing complex global features. In our framework, the DNN serves as the Spotter, identifying the rough global approximation, while the polynomial basis functions act as the Sniper, performing high-precision local refinements.

\subsection{From Nonlinear Optimization to Nonlinear Features Extraction and Linear Space Optimization}

\begin{figure}[H] 
    \centering
    \includegraphics[width=1.0\textwidth]{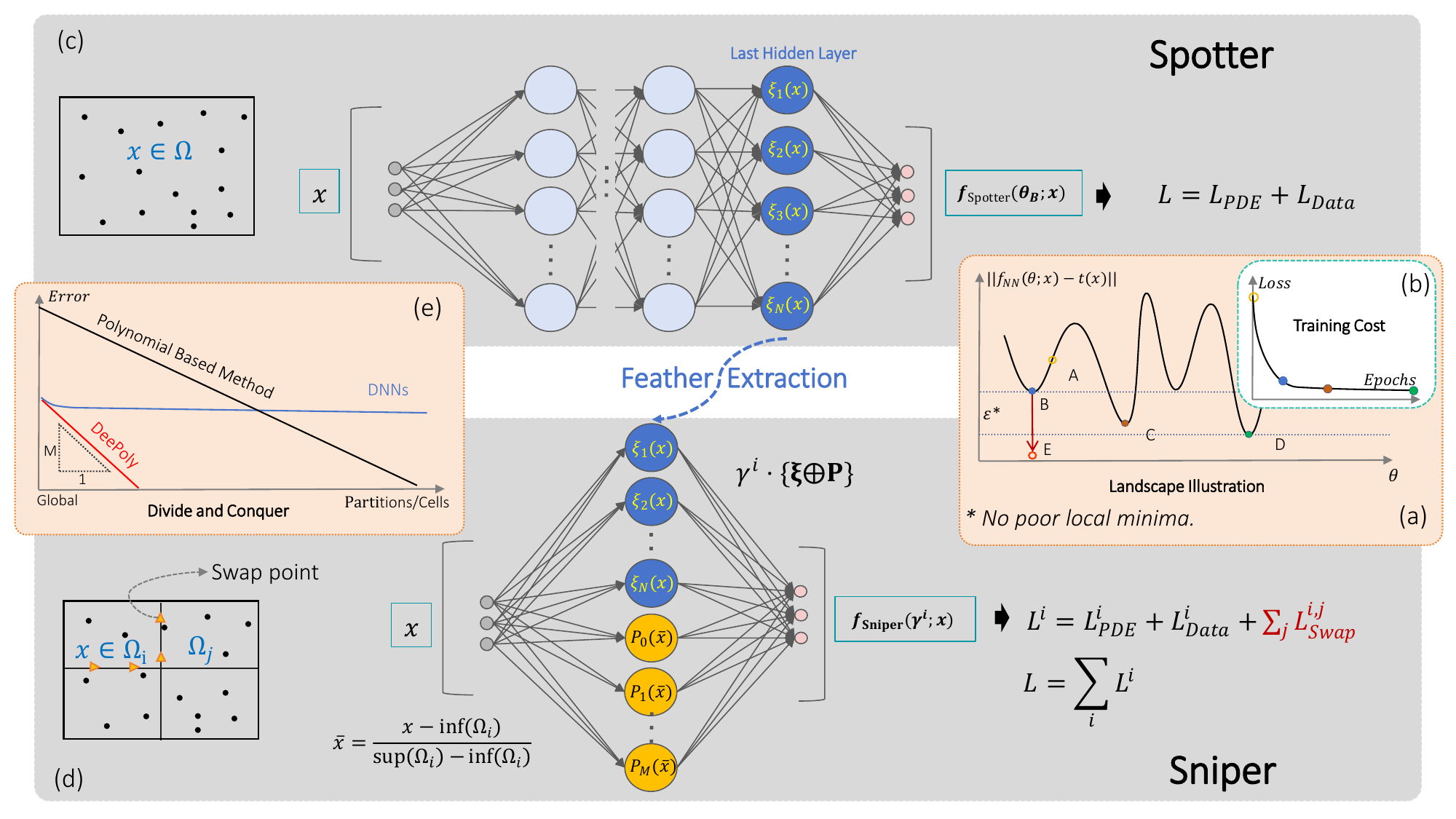} 
    \caption{Graphic illustration of DeePoly}
    \textbf{(a)} The local minimas are close to the global minimum however they are separated in parameter space.
    \textbf{(b)} It is effective to find a relatively good local minima however it expensive to traverse from one local minima to another.
    \textbf{(c)} Spotter works as a features extractor to obtain similar features of the target, without the requirement of high-precision approximation
    \textbf{(d)} Sniper works locally as a corrector to gain high-order accuracy.
    \textbf{(e)} Theoretically, DeePoly can achieve high-order (same to the order of the polynomial basis) accuracy by the guarantee of Taylor expansion of smooth function, and the requirement of partitions is lower than traditional methods.
    \label{fig:opt_landscape}
\end{figure}

Consider a typical non-convex optimization problem as illustrated in Figure \ref{fig:opt_landscape}.(a-b). When parameter optimization begins from point A, conventional approaches can readily find nearby local minima such as points B. However, locating smaller local minima or the global minimum remains exceptionally challenging and expensive.

Consequently, reducing the loss (or function error) from point B (e.g., \( 1 \times 10^{-3} \)) to point D (e.g., \( 1 \times 10^{-4} \)) or lower becomes extremely difficult through parameter optimization alone, even though \( f(\theta_B, x) \) and \( f(\theta_D, x) \) might represent functions with very similar features.

Based on these considerations, we propose shifting the optimization paradigm from exclusively optimizing parameters \( \theta \) to a two-step approach combining limited parameter optimization (\textbf{Spotter}) followed by linear space optimization (\textbf{Sniper}).

\textbf{Step 1 (Spotter):} Through parameter optimization, we find an acceptable solution with moderate error as illustrated as Fig.\ref{fig:opt_landscape}.(a-c) (for example, $Loss = 10^{-3}$), denoted as \( f_{\rm Spotter}(\theta_B; x) \). At this point, we cease parameter optimization, avoiding the diminishing returns associated with pursuing lower local minima through non-convex optimization.

\textbf{Step 2 (Sniper):} We extract the network features from the trained network at point B as approximate global features:
\[
f_{\rm Spotter}(\theta_B; x) = \sum_{n=1}^{N} \beta_n \xi_n(x), \quad x \in \Omega
\]
where \( \beta_n \) are the linear coefficients of the output layer, and \( \xi_n(x) \) represents the output of the last hidden layer, signifying the features extracted by the network for the target function. We define the space spanned by these features as:
\[
\boldsymbol{\xi} = \text{span}\{ \xi_1(x), \dots, \xi_N(x) \}.
\]
Then we supplement these features with polynomial basis functions to gain high-order accuracy with partition (as elaborated in the Theoretical section \ref{sec:theoretical}):
\[
\boldsymbol{P} = \text{span} \{ P_0(\bar{x}), P_1(\bar{x}), \dots, P_M(\bar{x}) \}
\]
where \( P_i(\bar{x}) \) are polynomial basis functions and \(\bar{x}\in \Omega_j\) is the partition of the domain \(\Omega\),and
\[
\bar{x} = \frac{x-\inf(\Omega_j)}{\sup(\Omega_j)-\inf(\Omega_j)}
\]
Then we construct the combined linear space:
\[
\mathbf{V} = \boldsymbol{\xi} \oplus \boldsymbol{P}
\]
 and perform linear optimization within the space \( \mathbf{V} \) to find the best approximation, aiming to reach point E:
\[
\min_{\gamma_n^i \in \mathbb{R}} \left\| \sum_{n=1}^K \gamma_n^i v_n(x) - t(x) \right\| \quad \text{where} \quad v_n(x) \in \mathbf{V}, K = \dim(\mathbf{V}) \quad x \in \Omega_i
\]
where \( t(x) \) is the target function.

Since continuous low-order polynomial features possess controllable spatial convergence properties (see Theoretical section), if point E does not meet the required accuracy, we can employ domain partitioning. The \textbf{Sniper} optimization is then performed on the combination of global features and local M-order features within smaller subdomains, potentially reaching machine precision theoretically.

\subsection{Solving Time-Dependent Nonlinear PDEs}

With the assistance of automatic differentiation, the solution process for PDEs follows a similar approach to function approximation, with special consideration given to the linearization of non-linear terms. While the optimization process described above is effective for directly solving non-linear functions-converting from fine-tuning network coefficients to fine-tuning fewer linear coefficients-we recommend linearizing non-linear terms to guarantee accuracy, given the challenges of non-linear optimization. Time discretization and pseudo-time stepping represent two effective approaches for this linearization.

We primarily develop two methods for solving equations. The first targets time-dependent equations using a time-stepping approach. Consider the equation:
\[
\frac{\partial U}{\partial t} = \mathbb{L}(U) + \mathbb{N}(U)\label{eq:time_dependent_pde}
\]
where \( \mathbb{L} \) is a linear spatial operator and \( \mathbb{N} \) is a non-linear spatial operator, both potentially involving various spatial derivatives. For simplicity, we illustrate using a first-order implicit time-stepping scheme, although higher-order schemes are implemented in the actual code:
\[
U^{n+1} - U^n = \Delta t^n \mathbb{L}(U^{n+1}) + \Delta t^n \mathbb{N}(U^n, U^{n+1}),
\]
For numerical stability, the \( \mathbb{N} \) term is linearized while minimizing dependency on \( U^n \). For example:
\[
U \frac{\partial U}{\partial x} \rightarrow U^n \frac{\partial U^{n+1}}{\partial x}
\]
For the linearized discrete equation, the solution process follows the same workflow as function fitting. All spatial derivatives are computed using automatic differentiation. Thus, the parameter optimization step (Step 1) employs the exact same procedure as PINNs. After completing the initial DNN training, we construct the linear space \( \mathbf{V} \) and then solve a system of linear equations (Step 2).

For problems involving multiple subdomains, continuity conditions must be imposed at the partition boundaries. We enforce continuity up to one order less than the highest spatial derivative order of \( U \) in the equation. For example, if the equation contains \( U_{xxx} \), continuity conditions for \( U, U_x, U_{xx} \) are required.

Between time steps, since \( U \) changes relatively little, the network does not require retraining. Only a transfer step is needed, combined with the fact that the required accuracy for network training is relatively low, resulting in a highly efficient solution process.

\subsection{Solving General PDEs}

Although time discretization effectively reduces the non-linearity and problem complexity of each step, even with high-order time integration schemes, temporal discretization still introduces non-negligible errors. For applications requiring minimal error introduction, global optimization approaches are preferable.

For steady-state equations, or when treating time as another spatial dimension solved via automatic differentiation, a pseudo-time stepping iterative method can be employed. For the equation \ref{eq:time_dependent_pde}, the scheme becomes:
\[
U^{n+1} - U^{n} = \Delta \tau \left( \frac{\partial U^{n+1}}{\partial t} - \mathbb{L}(U^{n+1}) - \mathbb{N}(U^{n+1}, U^n) \right)
\]
Here, the initial condition \( U^0 \) can be a uniform field. A more efficient convergence strategy involves using the solution \( U(t, x) \) obtained from a preliminary PINN solution of the original global problem as the initial field. Subsequently, no further network training is performed; only iterative solving of linear systems is needed to optimize the coefficients of the basis functions in the space \( \mathbf{V} \). Since the required accuracy for the initial field is low, and subsequent steps avoid network training, the solution process remains highly efficient.

\section{Numerical Examples}
In all examples, the sampling points (for training) are generated using uniform random sampling, demonstrating the mesh-free characteristic of the new method. The GPU used is a Titan V, which helps in evaluating the computation time cross-device.

\subsection{Function Approximation}

First, we test the approximation performance on smooth functions, considering one-dimensional and two-dimensional examples:

1D function:
\[
t_1(x) = \sin(2 \pi x) + 0.5 \cos(4 \pi x)
\]
(Note: Assuming the code snippet for the 2D function definition in the markdown was meant to define the target function, but it wasn't explicitly assigned to a variable. Translating the calculation.)

2D function calculation described:
\[
  \left\{
\begin{aligned}
t_2(x,y) &= g_1(x,y) + g_2(x,y) + g_3(x,y) + s(x,y) + p(x,y) \\
g_1(x,y) &= \exp(-(x^2 + y^2)) \\
g_2(x,y) &= 0.7 \exp\left(-\frac{(x-1.5)^2 + (y-1.0)^2}{1.2}\right) \\
g_3(x,y) &= 0.5 \exp\left(-\frac{(x+1.0)^2 + (y+1.5)^2}{0.8}\right) \\
s(x,y) &= 0.2 \sin(3x) \cos(2y) \\
p(x,y) &= 0.1 (x^2 - y^2) + 0.05 x y^2
\end{aligned}
\right.
\]

\begin{table}[H]
    \centering
    \caption{Approximation results for the 1D smooth function.}
    \label{tab:smooth_1d}
    \small
    \begin{tabular}{@{}cc|c|ccc|ccc|c@{}}
        \toprule
        \multirow{2}{*}{Points} & \multirow{2}{*}{Sections} & \multirow{2}{*}{Loss} & \multicolumn{3}{c|}{Training} & \multicolumn{3}{c|}{Testing} & \multirow{2}{*}{Time(s)} \\ 
        \cmidrule(lr){4-6} \cmidrule(lr){7-9}
        & & & MSE & MAE & Order & MSE & MAE & Order & \\
        \midrule
        200   & 1  & 2.1e-6 & 1.2e-06 & 1.5e-03 & -- & 1.6e-06 & 1.3e-03 & -- & 7.1 \\ 
        2000  & 1  & 8.6e-07 & 5.1e-07 & 5.9e-04 & -- & 5.1e-07 & 5.9e-04 & -- & 10.7 \\ 
        2000  & 2  & 2.4e-06 & 1.2e-13 & 2.6e-07 & 11.2 & 1.2e-13 & 2.7e-07 & 11.1 & 9.9 \\ 
        2000  & 4  & 1.6e-06 & 8.4e-22 & 2.1e-11 & 13.6 & 9.9e-22 & 2.3e-11 & 13.5 & 9.7 \\ 
        2000  & 8  & 2.2e-06 & 9.5e-25 & 5.2e-13 & 5.3 & 1.0e-23 & 7.2e-13 & 5.0 & 9.1 \\ 
        2000  & 16 & 1.0e-06 & 2.0e-26 & 8.2e-14 & 2.7 & 5.8e-25 & 1.5e-13 & 2.3 & 11.9 \\ 
        2000  & 32 & 3.8e-06 & 8.8e-27 & 3.4e-14 & 1.3 & 3.3e-26 & 5.8e-14 & 1.4 & 9.6 \\ 
        \bottomrule
    \end{tabular}
    \caption*{This example employs 10th-order polynomial basis functions, which explains the approximately 10th-order convergence accuracy (11-13) observed in the initial refinements. The convergence order naturally decreases as the solution approaches machine precision.}
\end{table}

\begin{table}[H]
    \centering
    \caption{Approximation results for the 2D smooth function.}
    \label{tab:smooth_2d}
    \small
    \begin{tabular}{@{}cc|c|ccc|ccc|c@{}}
        \toprule
        \multirow{2}{*}{Points} & \multirow{2}{*}{Sections} & \multirow{2}{*}{Loss} & \multicolumn{3}{c|}{Training} & \multicolumn{3}{c|}{Testing} & \multirow{2}{*}{Time(s)} \\ 
        \cmidrule(lr){4-6} \cmidrule(lr){7-9}
        & & & MSE & MAE & Order & MSE & MAE & Order & \\
        \midrule
        20000 & $1^2$ & 2.9e-05 & 2.3e-05 & 3.6e-03 & -- & 2.8e-05 & 3.8e-03 & -- & 36.5 \\ 
        20000 & $2^2$ & 3.3e-05 & 2.3e-06 & 1.2e-03 & 1.6 & 3.1e-06 & 1.3e-03 & 1.6 & 38.7 \\ 
        20000 & $4^2$ & 3.2e-05 & 6.1e-11 & 5.1e-06 & 7.8 & 1.1e-10 & 5.8e-06 & 7.7 & 38.9 \\ 
        20000 & $6^2$ & 2.8e-05 & 2.9e-14 & 1.1e-07 & 9.5 & 1.6e-13 & 1.4e-07 & 9.1 & 40.2 \\ 
        20000 & $8^2$ & 2.9e-05 & 8.5e-17 & 5.1e-09 & 8.1 & 9.9e-16 & 9.7e-09 & 7.2 & 42.2 \\ 
        20000 & $10^2$ & 2.9e-05 & 6.8e-19 & 4.0e-10 & 6.4 & 4.5e-17 & 1.2e-09 & 5.2 & 47.6 \\ 
        \bottomrule
    \end{tabular}
    \caption*{This example employs 5th-order polynomial basis functions in each dimension, resulting in high convergence orders in the 2D case.}
\end{table}

The smooth function examples are intended to assess whether the new method exhibits convergence accuracy and whether the solution time remains manageable. Next, we consider its performance on problems with discontinuities or large gradients.
\begin{figure}[H]
    \centering
    \includegraphics[width=0.8\textwidth]{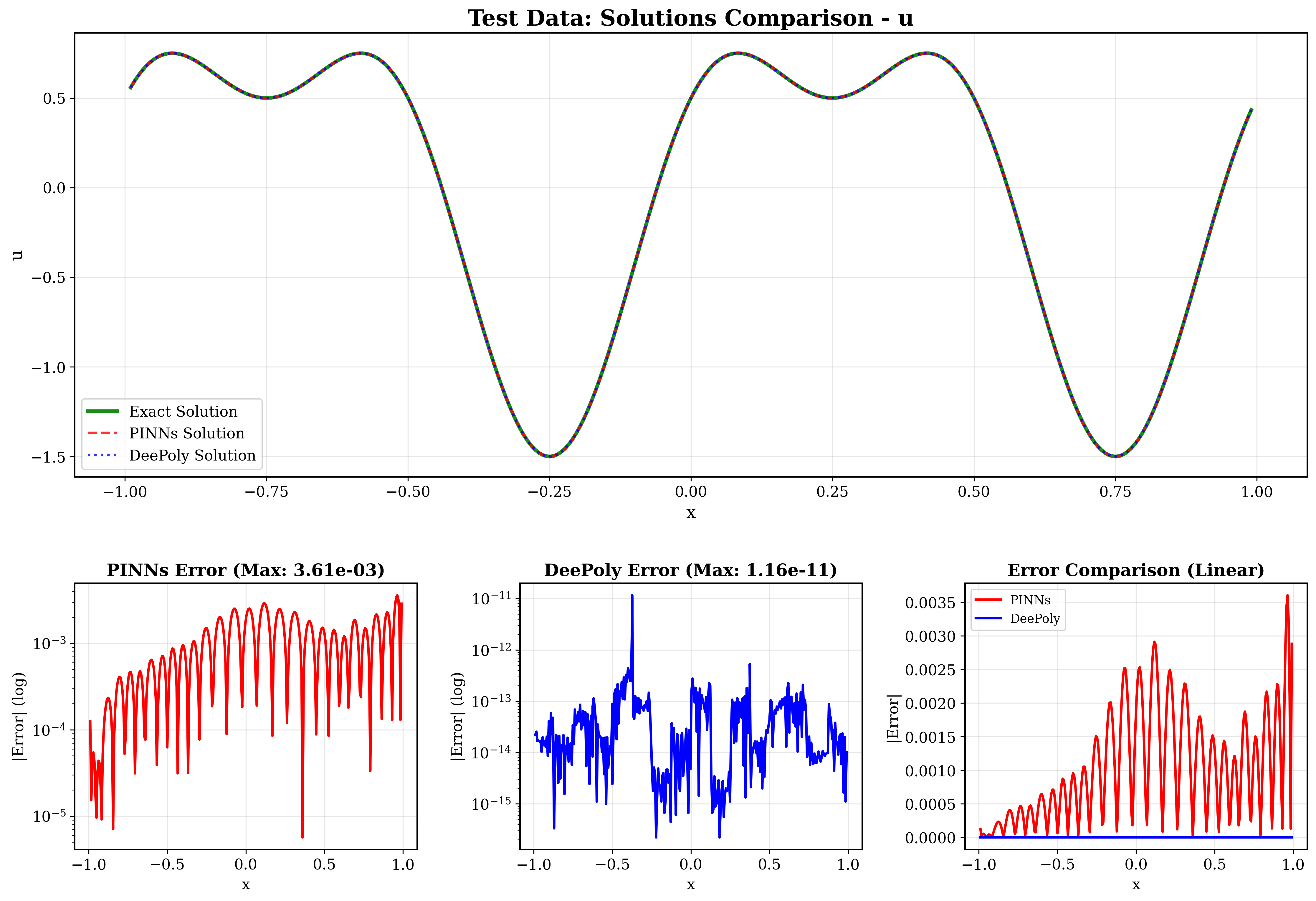}
    \caption{1D smooth function example result with 32 partitions.}
    \label{fig:smooth_1d_res}
\end{figure}

\begin{figure}[H]
    \centering
    \includegraphics[width=0.8\textwidth]{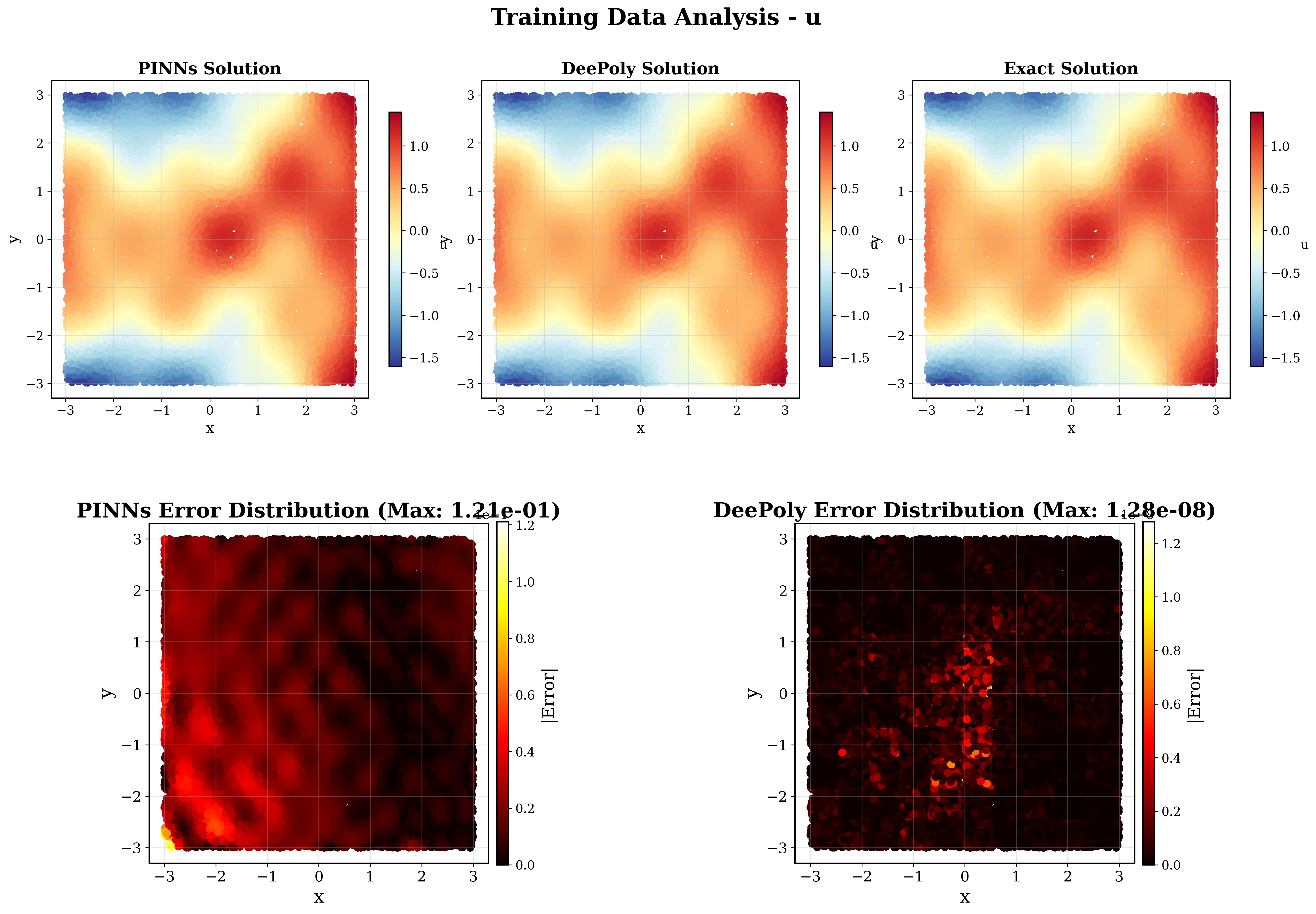}
    \includegraphics[width=0.8\textwidth]{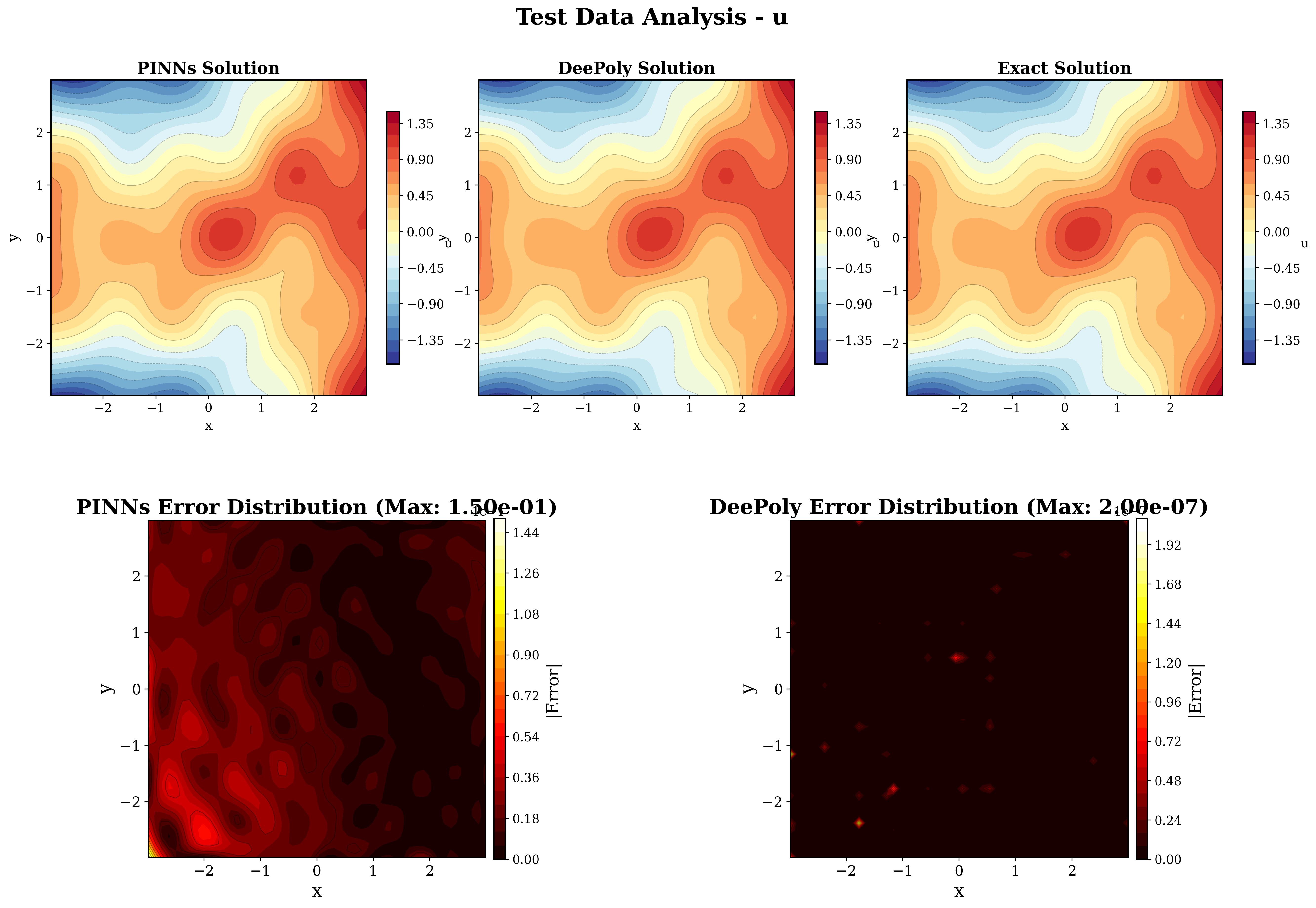}
    \caption{2D smooth function example result with 10x10 partitions.}
    \label{fig:smooth_2d_res}
\end{figure}

\subsection{Discontinuous/Large Gradient Function Approximation}

Considering that polynomial functions cannot stably approximate large gradients and discontinuities, we test the new method's capability to capture complex features and its accuracy in such scenarios.

The 1D problem is defined as:
\[
y(x) = 
\begin{cases} 
      5 + \sum_{k=1}^{4} \sin(kx) & x < 0 \\\\ 
      \cos(10x) & x \geq 0 
\end{cases}
\]
\begin{table}[H]
    \centering
    \caption{Approximation results for the 1D discontinuous function.}
    \label{tab:discont_1d}
    \small
    \begin{tabular}{@{}cc|c|ccc|ccc|c@{}}
        \toprule
        \multirow{2}{*}{Nodes} & \multirow{2}{*}{Partitions} & \multirow{2}{*}{Loss} & \multicolumn{3}{c|}{Training} & \multicolumn{3}{c|}{Testing} & \multirow{2}{*}{Time(s)} \\ 
        \cmidrule(lr){4-6} \cmidrule(lr){7-9}
        & & & MSE & MAE & Order & MSE & MAE & Order & \\
        \midrule
        2000 & 1  & 2.0e-05 & 2.0e-05 & 3.1e-03 & -- & 2.1e-05 & 3.2e-03 & -- & 18.2 \\ 
        2000 & 3  & 2.0e-05 & 4.8e-06 & 1.3e-03 & 0.9 & 5.2e-06 & 1.3e-03 & 0.9 & 17.6 \\ 
        2000 & 9  & 2.0e-05 & 1.3e-09 & 7.3e-06 & 4.5 & 1.6e-08 & 1.5e-05 & 4.0 & 18.2 \\ 
        2000 & 27 & 2.0e-05 & 7.5e-12 & 4.2e-07 & 2.6 & 1.4e-10 & 1.4e-06 & 2.1 & 17.9 \\ 
        \bottomrule
    \end{tabular}
    \caption*{This example demonstrates the method's ability to handle discontinuities. The convergence orders for both training and testing are high despite the function's discontinuity, showing that the DNN captures the global features while polynomial bases handle local refinements.}
\end{table}

The results show that despite using polynomial bases, oscillation or stability issues did not arise, indicating that the DNN captured the discontinuity features in the discontinuous region and can stabilize potential oscillations caused by polynomial bases.

\begin{figure}[H]
    \centering
    \includegraphics[width=0.8\textwidth]{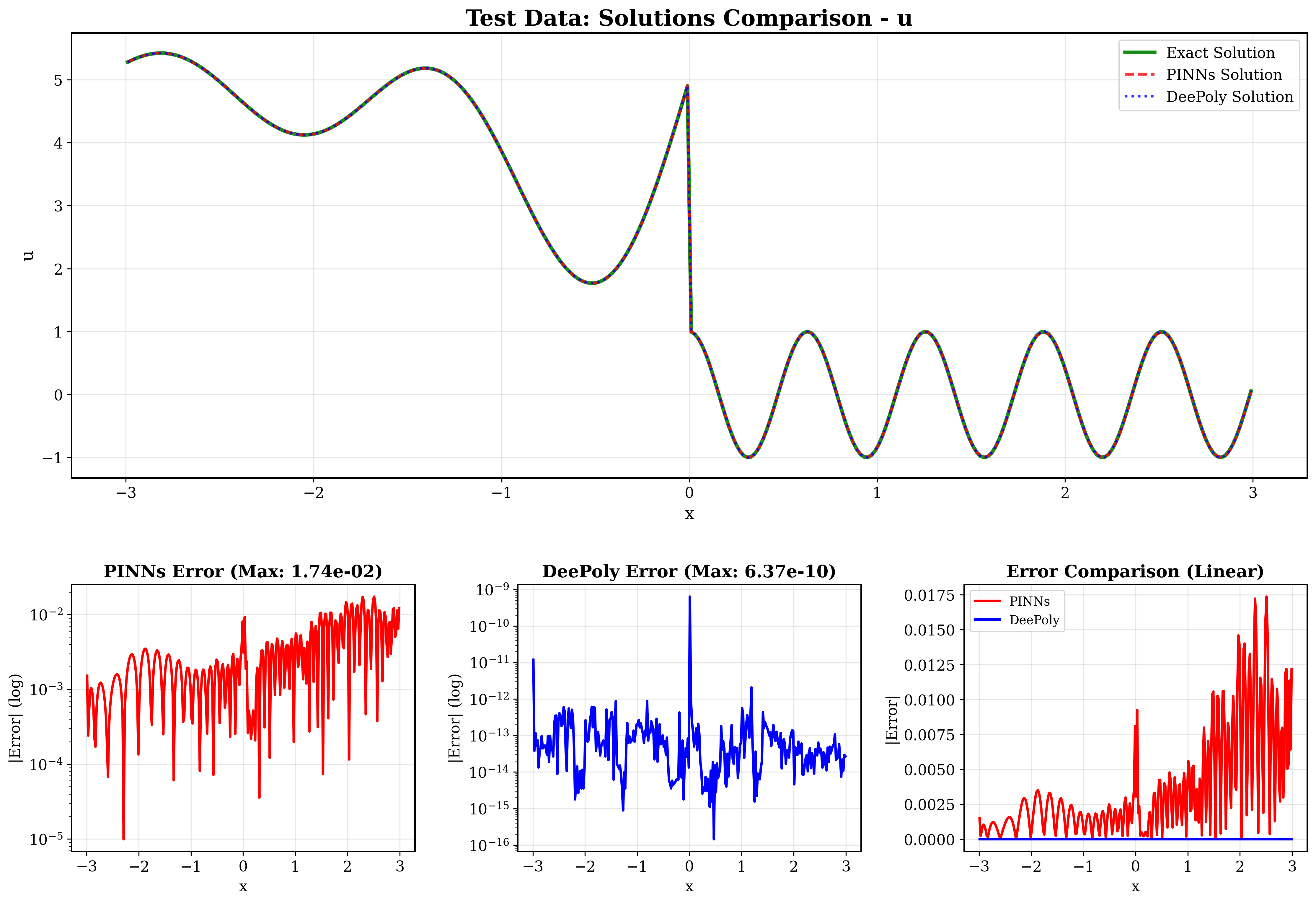}
    \caption{1D discontinuous function result.}
    \label{fig:discont_1d_1}
\end{figure}

Next, we test a 2D example with complex large gradients. The target function is defined as:

\[
\left\{
\begin{aligned}
f(x,y) &= z(x,y) + g(x,y) \\
z(x,y) &= h \tanh(s \cdot d(x,y)) \\
d(x,y) &= y - A \sin(f \pi x) \\
g(x,y) &= 0.4 \sin(3x) \cos(2y)
\end{aligned}
\right.
\]

with parameters are set as:
\[ A = 0.4, f = 2.0, s = 15.0, h = 8.0 \]

\begin{table}[H]
    \centering
    \caption{Approximation results for the 2D large gradient function.}
    \label{tab:discont_2d}
    \small
    \begin{tabular}{@{}cc|c|ccc|ccc|c@{}}
        \toprule
        \multirow{2}{*}{Points} & \multirow{2}{*}{Sections} & \multirow{2}{*}{Loss} & \multicolumn{3}{c|}{Training} & \multicolumn{3}{c|}{Testing} & \multirow{2}{*}{Time(s)} \\ 
        \cmidrule(lr){4-6} \cmidrule(lr){7-9}
        & & & MSE & MAE & Order & MSE & MAE & Order & \\
        \midrule
        20000 & $1^2$  & 2.0e-05 & 1.7e-05 & 2.9e-03 & -- & 1.8e-05 & 3.0e-03 & -- & 47.0 \\ 
        20000 & $2^2$  & 2.0e-05 & 1.1e-05 & 2.0e-03 & 0.6 & 1.2e-05 & 2.0e-03 & 0.6 & 46.9 \\ 
        20000 & $4^2$  & 2.0e-05 & 4.1e-06 & 7.7e-04 & 1.4 & 4.9e-06 & 7.9e-04 & 1.3 & 47.1 \\ 
        20000 & $8^2$  & 2.0e-05 & 1.5e-08 & 4.1e-05 & 4.2 & 1.6e-07 & 6.9e-05 & 3.5 & 49.2 \\ 
        20000 & $10^2$ & 2.0e-05 & 9.9e-10 & 9.5e-06 & 3.0 & 7.6e-08 & 2.3e-05 & 2.2 & 51.6 \\ 
        \bottomrule
    \end{tabular}
    \caption*{This example shows the method's effectiveness on functions with large gradients. The high convergence orders demonstrate how the combined approach handles complex gradient structures.}
\end{table}

\begin{figure}[H]
    \centering
    \includegraphics[width=0.8\textwidth]{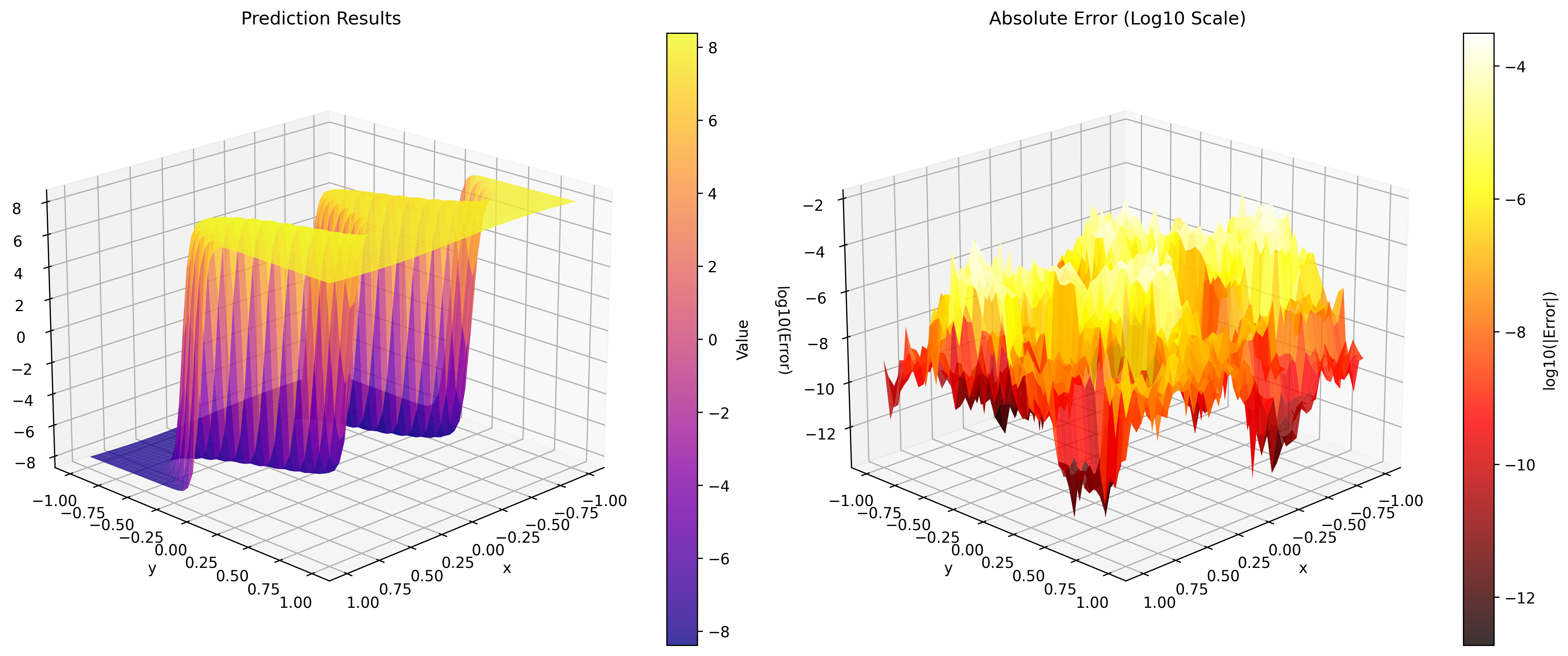}
    \caption{2D large gradient problem result (10x10 partitions).}
    \label{fig:discont_2d_1010}
\end{figure}

\subsection{Linear PDEs}

\subsubsection{Poisson Equation}
Here we consider the Poisson equation with Dirichlet boundary conditions:

\[
\begin{aligned}
\nabla^2 u &= f(x,y) \\
u &= g(x,y) \quad \text{on} \quad \partial \Omega
\end{aligned}
\]
where we test two different source functions to assess the method's performance on problems with varying frequency content:

\begin{align}
\text{Case 1: } \quad f(x,y) &= -2\pi^2\sin(\pi x) \cos(\pi y) \\
\text{Case 2: } \quad f(x,y) &= -32\pi^2\sin(4\pi x) \cos(4\pi y)
\end{align}
The corresponding analytical solutions are:
\begin{align}
\text{Case 1: } \quad u(x,y) &= \sin(\pi x) \cos(\pi y) \\
\text{Case 2: } \quad u(x,y) &= \sin(4\pi x) \cos(4\pi y)
\end{align}

The basic settings are:

The computational configuration for both cases is:
\begin{itemize}
    \item Training Points: 5000
    \item Partitions: $10 \times 10$
    \item Polynomial Degree: $[5, 5]$
    \item Hidden Dimensions: $[12, 32, 32, 25]$
    \item Domain: $[0, 1]^2$
    \item Test Grid: $50 \times 50$
\end{itemize}
\textbf{Case 1: } $f(x,y) = -2\pi^2\sin(\pi x)\cos(\pi y)$

\begin{table}[H]
    \centering
    \caption{Performance comparison for 2D Poisson equation (Case 1)}
    \label{tab:poisson_case1}
    \small
    \begin{tabular}{@{}lccccc@{}}
        \toprule
        Method & Dataset & MSE & MAE & Max Error & Time (s) \\
        \midrule
        PINNs & Train & 1.04e-06 & 8.51e-04 & 4.50e-03 & 13.73 \\
        DeePoly & Train & 5.57e-32 & 1.75e-16 & 2.37e-15 & 18.06 \\
        \midrule
        PINNs & Test & 1.06e-06 & 8.58e-04 & 4.54e-03 & -- \\
        DeePoly & Test & 6.11e-32 & 1.81e-16 & 2.24e-15 & -- \\
        \bottomrule
    \end{tabular}
\end{table}

\textbf{Case 2: } $f(x,y) = -32\pi^2\sin(4\pi x)\cos(4\pi y)$

\begin{table}[H]
    \centering
    \caption{Performance comparison for 2D Poisson equation (Case 2)}
    \label{tab:poisson_case2}
    \small
    \begin{tabular}{@{}lccccc@{}}
        \toprule
        Method & Dataset & MSE & MAE & Max Error & Time (s) \\
        \midrule
        PINNs & Train & 1.27e-07 & 2.60e-04 & 1.89e-03 & 34.73 \\
        DeePoly & Train & 5.55e-19 & 5.79e-10 & 7.67e-09 & 49.10 \\
        \midrule
        PINNs & Test & 1.29e-07 & 2.62e-04 & 1.74e-03 & -- \\
        DeePoly & Test & 5.43e-19 & 5.86e-10 & 4.18e-09 & -- \\
        \bottomrule
    \end{tabular}
\end{table}

\begin{figure}[H]
    \centering
    \includegraphics[width=0.8\textwidth]{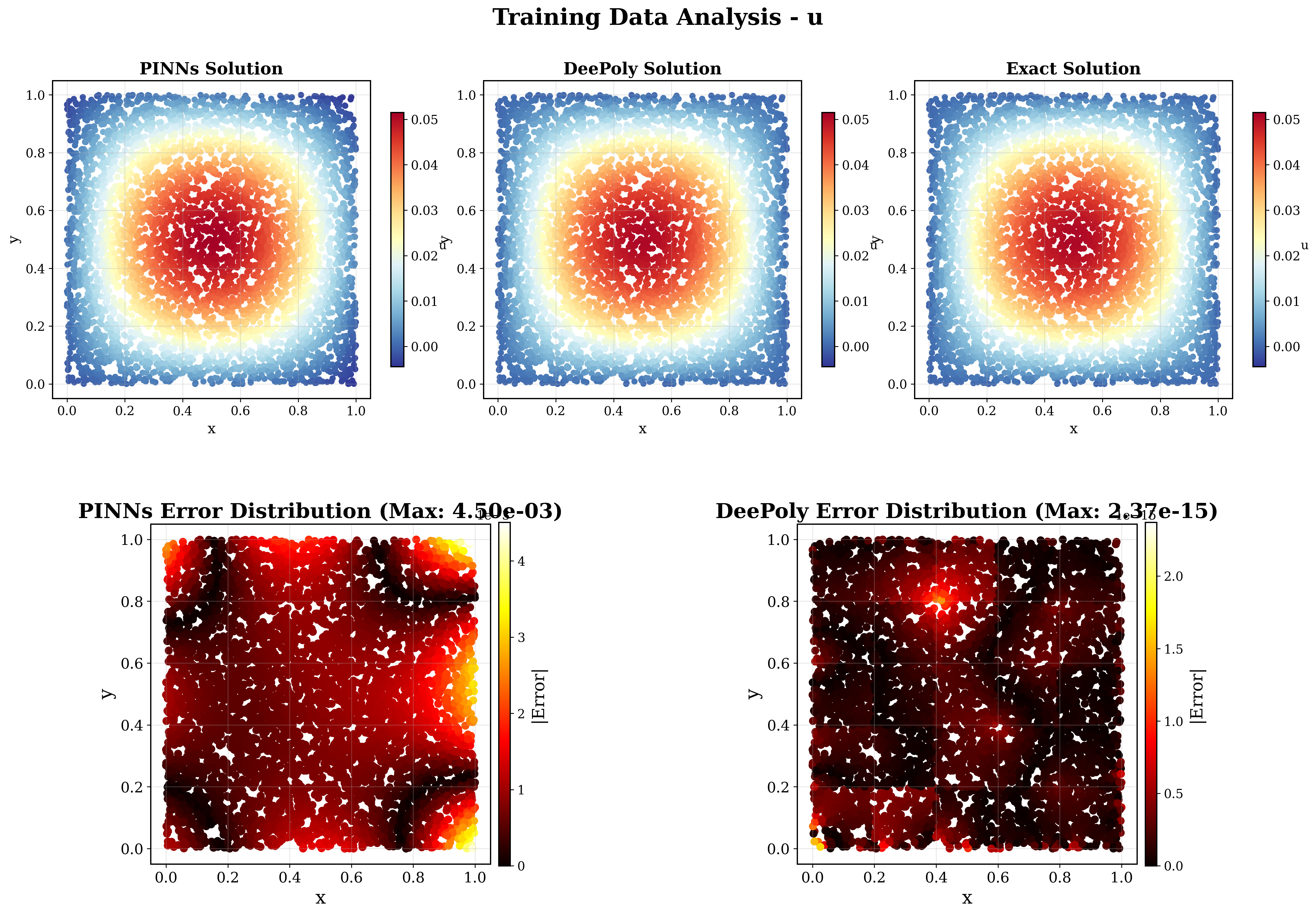}
    \includegraphics[width=0.8\textwidth]{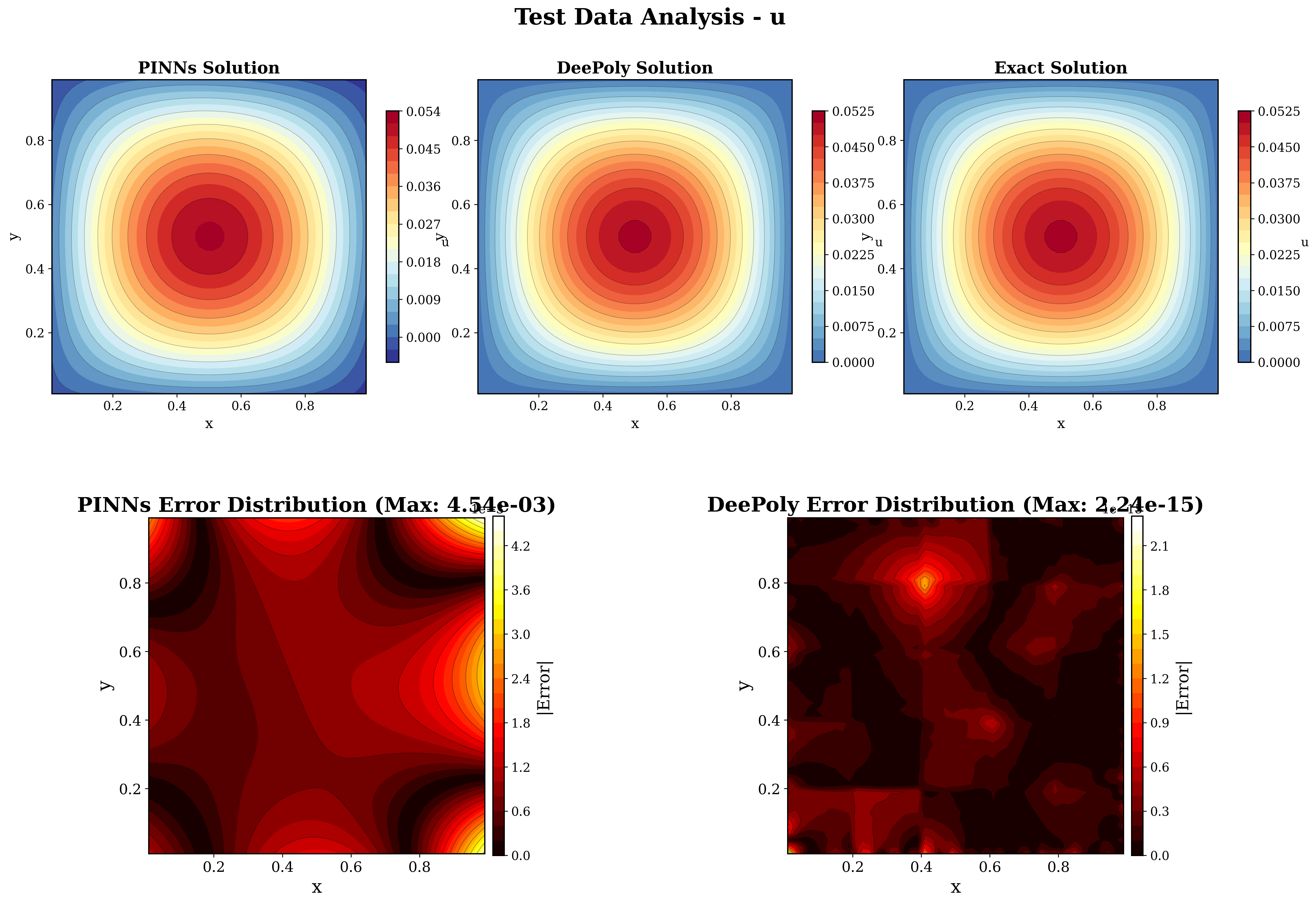}
    \caption{2D Poisson equation with $\sin(\pi x)\sin(\pi y)$ source term }
    \label{fig:poisson_sinpixsinpiy}
\end{figure}

\begin{figure}[H]
    \centering
    \includegraphics[width=0.8\textwidth]{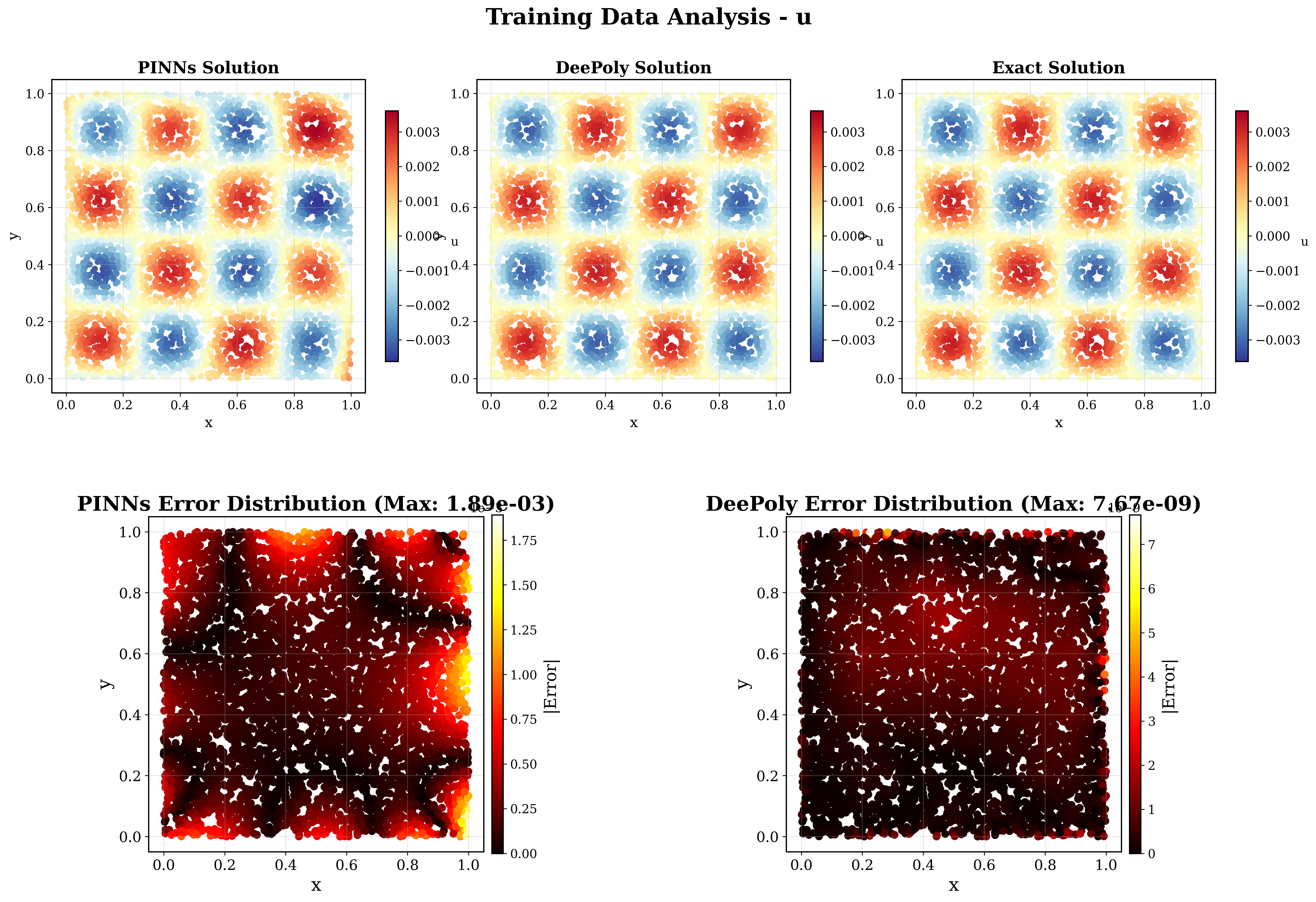}
    \includegraphics[width=0.8\textwidth]{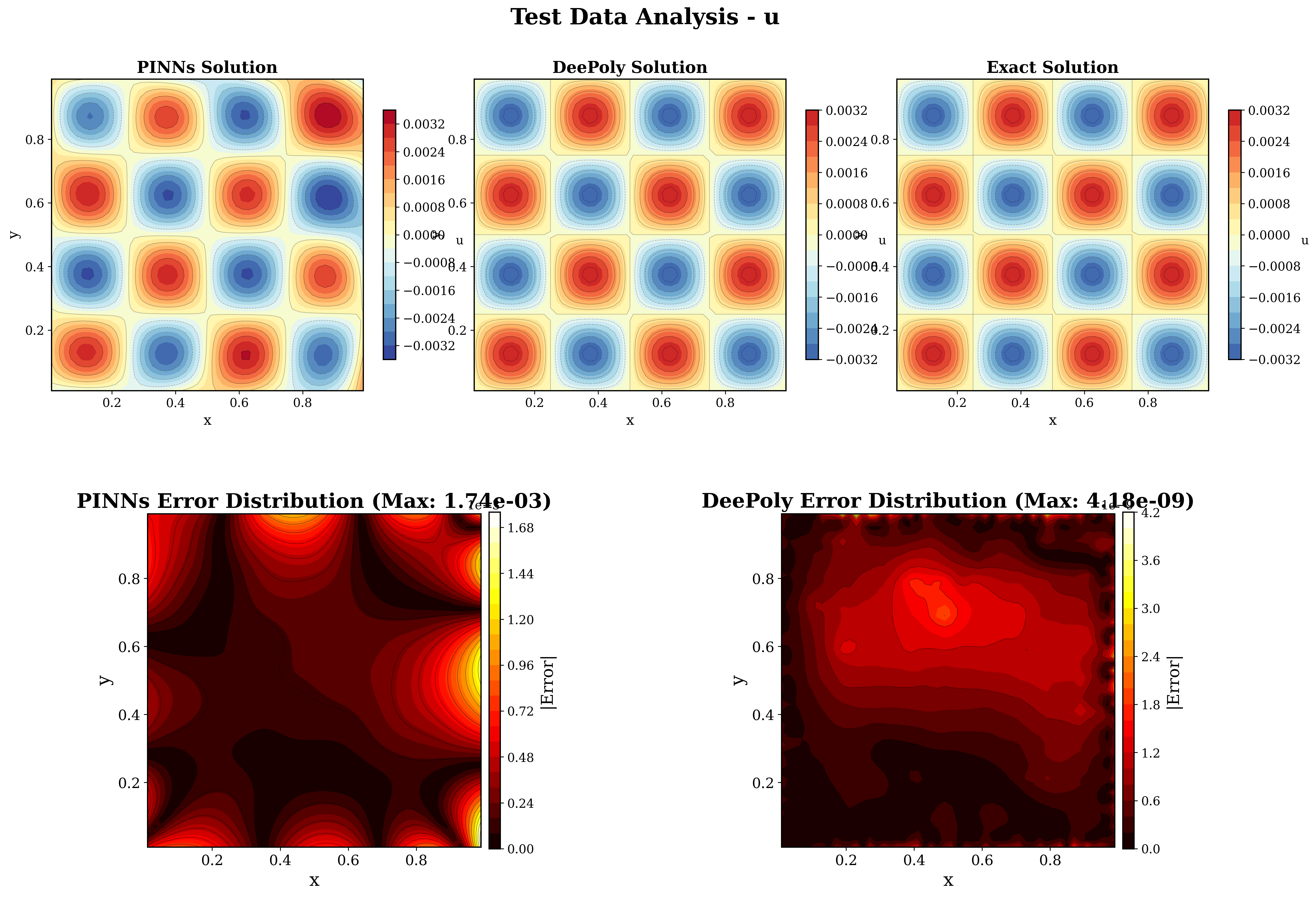}
    \caption{2D Poisson equation with $\sin(4\pi x)\sin(4\pi y)$ source term }
    \label{fig:poisson_sin4pixsin4piy}
\end{figure}

\subsubsection{Linear Convection Equation}

Then we consider the linear convection equation with a large gradient initial condition:

\[
\begin{aligned}
\frac{\partial u}{\partial t} + a \frac{\partial u}{\partial x} = 0 \\
u(x, 0) = \tanh(100*(x-0.3))
\end{aligned}
\]

where $a = 0.3$ is the convection speed. The initial condition is a large gradient function, which is challenging for numerical methods, especially may cause instability if without the consideration of upwind properties in the construction of the numerical scheme.

The computational configuration is:
\begin{itemize}
    \item Training Points: 3000
    \item Partitions: $1 \times 20$ (temporal $\times$ spatial)
    \item Polynomial Degree: $[5, 5]$
    \item Hidden Dimensions: $[12, 32, 32, 25]$
    \item Domain: $[0, 1]^2$ (time $\times$ space)
    \item Test Grid: $50 \times 50$
\end{itemize}

\begin{table}[H]
    \centering
    \caption{Performance comparison for linear convection equation with discontinuous initial condition}
    \label{tab:linear_convection}
    \small
    \begin{tabular}{@{}lccccc@{}}
        \toprule
        Method & Dataset & MSE & MAE & Max Error & Time (s) \\
        \midrule
        PINNs & Train & 4.47e-07 & 3.55e-04 & 4.01e-03 & 16.83 \\
        DeePoly & Train & 3.69e-08 & 4.97e-05 & 2.11e-03 & 17.45 \\
        \midrule
        PINNs & Test & 4.16e-07 & 3.39e-04 & 4.00e-03 & -- \\
        DeePoly & Test & 3.24e-08 & 4.59e-05 & 2.08e-03 & -- \\
        \bottomrule
    \end{tabular}
\end{table}

\begin{figure}[H]
    \centering
    \includegraphics[width=0.8\textwidth]{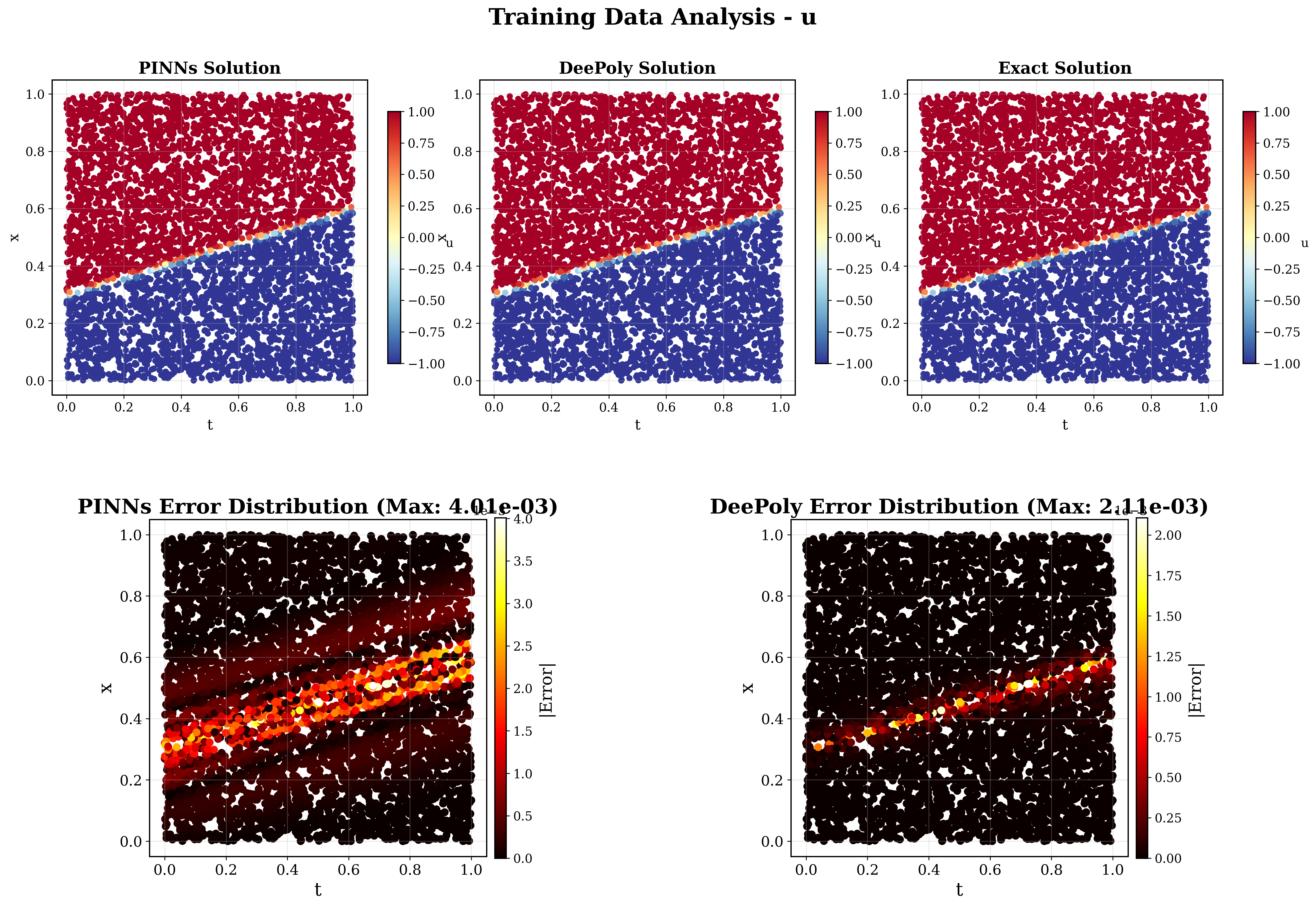}
    \includegraphics[width=0.8\textwidth]{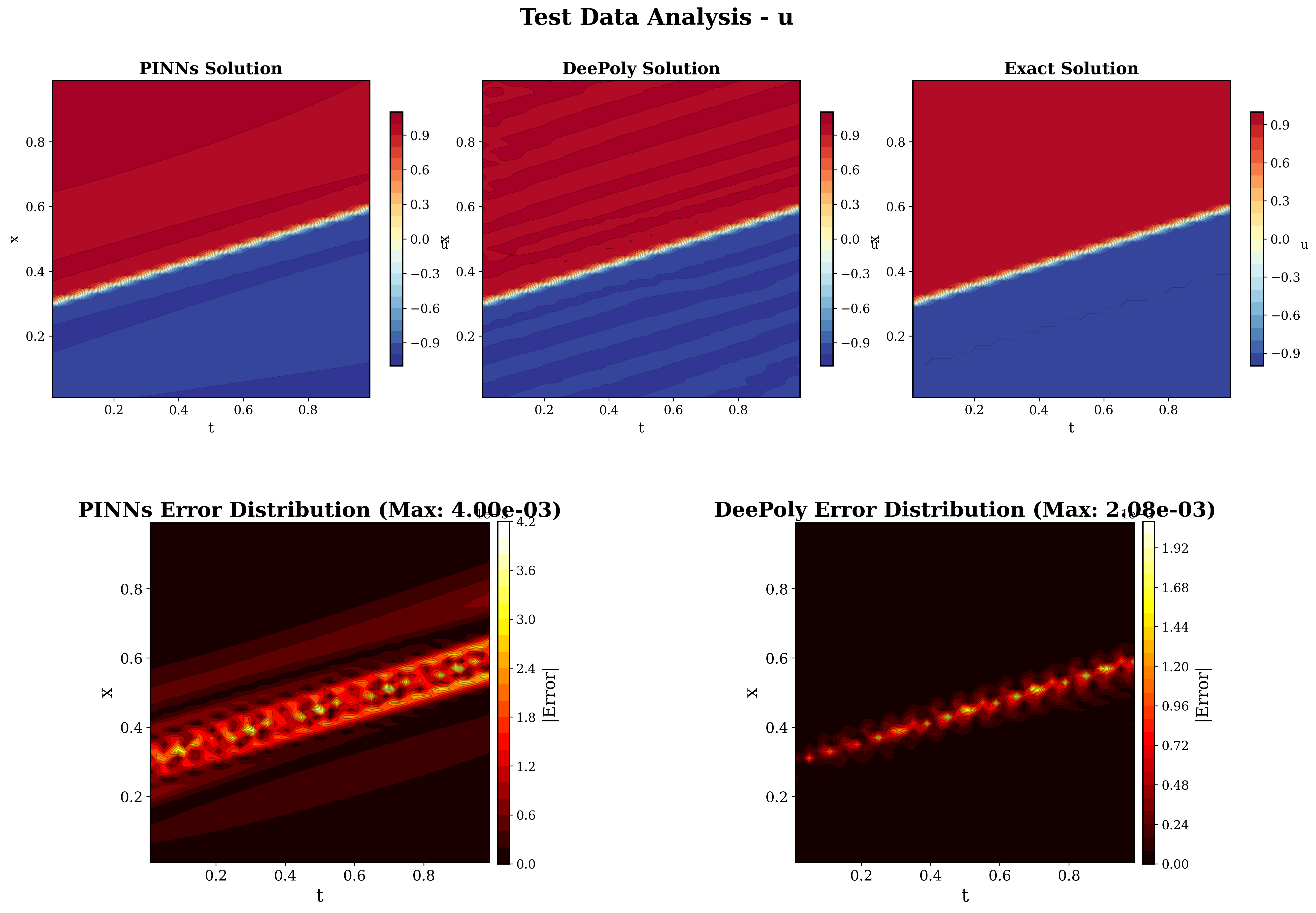}
    \caption{Linear convection equation with discontinuous initial condition. The method maintains stability and accuracy even in the presence of discontinuities.}
    \label{fig:linear_convection_discontinuity}
\end{figure}

As shown in Figure \ref{fig:linear_convection_discontinuity}, the method successfully captures the large gradient initial condition and maintains stability in the presence of discontinuities without requiring explicit upwind schemes. This remarkable stability is primarily attributed to the global perspective of PINNs and the synergistic interaction between neural networks and polynomial basis functions. Specifically, the DNNs serve as stabilizers for the polynomial approximation, while the polynomial basis functions effectively compensate for the local accuracy deficiencies inherent in DNNs. This validates the powerful combination of the Spotter+Sniper framework.

It should be noted that while the current accuracy for discontinuous problems does not reach the machine precision levels achieved for smooth problems, this limitation stems from the conditioning properties of the large linear system that must be solved in the presence of discontinuities. Future work will explore the integration of Physics-Informed Neural Networks with Weighted Essentials (PINNs-WE) \cite{pinn_we_reference} to effectively modulate the influence of discontinuous regions on the matrix conditioning, thereby further enhancing accuracy in smooth regions of discontinuous problems \cite{pinn_we_reference}.

\subsection{Nonlinear PDEs}
\subsubsection{Allen-Cahn Equation}

The Allen-Cahn equation is a nonlinear reaction-diffusion equation commonly used to model phase separation in multi-component alloy systems. In this example, we consider the following form:

\[
\frac{\partial u}{\partial t} = \epsilon^2 \frac{\partial^2 u}{\partial x^2} + u - u^3
\]

where $u(x,t)$ represents the order parameter with $x \in [-1, 1]$, $\epsilon = 0.01$ controls the interface width, and the nonlinear reaction term $u - u^3$ drives the phase separation. We impose the following initial and boundary conditions:

\[
\begin{aligned}
u(x, 0) &= x^2 \cos(\pi x)\\
u(-1, t) &= u(1, t) = -0.5
\end{aligned}
\]

The Allen-Cahn equation presents significant challenges due to its strong nonlinearity and the development of sharp transition layers as the solution evolves. Using our hybrid approach with the time-stepping method described in Section 2.2, we can efficiently solve this equation while maintaining high accuracy.

\begin{figure}[H]
    \centering
    \includegraphics[width=0.8\textwidth]{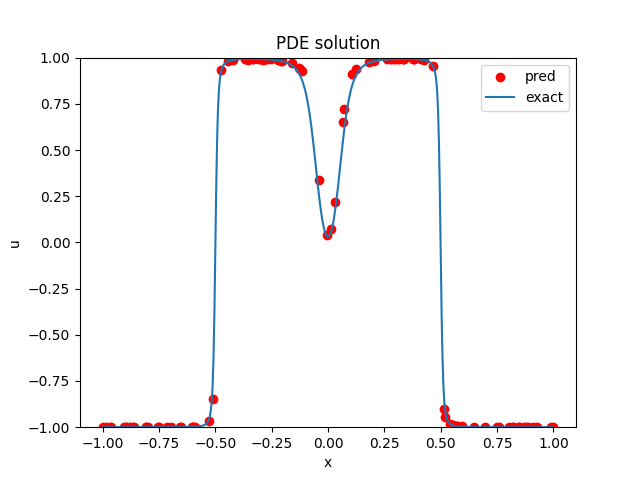}
    \caption{Allen-Cahn equation result at T=1.}
    \label{fig:ac_result}
\end{figure}

Figure \ref{fig:ac_result} shows the result using 115 random sampling points and 3 partitions. The time step is \( dt = 0.1 \), and the solution time is 22.5s. The figure indicates that the new method can achieve highly accurate results even with very sparse sampling points. The solution correctly captures the phase separation process and the sharp transition layers that form at T=1. Notably, traditional numerical methods would typically require significantly finer discretization to achieve comparable accuracy for this problem.

\subsubsection{Navier-Stokes Equation - Lid-Driven Cavity Flow}

The incompressible Navier-Stokes equations represent one of the most fundamental and challenging systems in fluid dynamics, described by:

\[
\begin{aligned}
\frac{\partial \mathbf{u}}{\partial t} + (\mathbf{u} \cdot \nabla)\mathbf{u} &= -\nabla p + \frac{1}{\text{Re}} \nabla^2 \mathbf{u} \\
\nabla \cdot \mathbf{u} &= 0
\end{aligned}
\]

where $\mathbf{u} = (u, v)$ is the velocity field, $p$ is the pressure, and $\text{Re}$ is the Reynolds number, which characterizes the ratio of inertial forces to viscous forces.

The lid-driven cavity flow is a classical benchmark problem in computational fluid dynamics. We consider a square domain $\Omega = [0,1]^2$ with no-slip boundary conditions on three walls (left, right, and bottom) while the top wall moves with a constant horizontal velocity. Specifically:
\[
\begin{aligned}
u(x, 0, t) &= u(0, y, t) = u(1, y, t) = 0 \\
v(x, 0, t) &= v(0, y, t) = v(1, y, t) = 0 \\
u(x, 1, t) &= 1, \quad v(x, 1, t) = 0 
\end{aligned}
\]

This problem features complex flow structures including main circulation and secondary vortices in the corners, with the flow complexity increasing at higher Reynolds numbers. We solve this problem at $\text{Re} = 100$ using the pseudo-time stepping approach described in Section 2.3, where we treat the steady-state Navier-Stokes equations as the target and iterate with a pseudo-time stepping scheme.

Our convergence criterion is based on the maximum norm of the velocity change between consecutive iterations:
\[
\|u^{n+1} - u^n\|_{L^\infty} + \|v^{n+1} - v^n\|_{L^\infty} \leq 10^{-4} 
\]

The solution process initializes with a preliminary approximation obtained from a global PINN solution with relatively low accuracy requirements. This approximation serves as the starting point for our hybrid approach, where we extract features from the neural network and combine them with polynomial basis functions for subsequent linear space optimization.

\begin{figure}[H]
    \centering
    \includegraphics[width=0.6\textwidth]{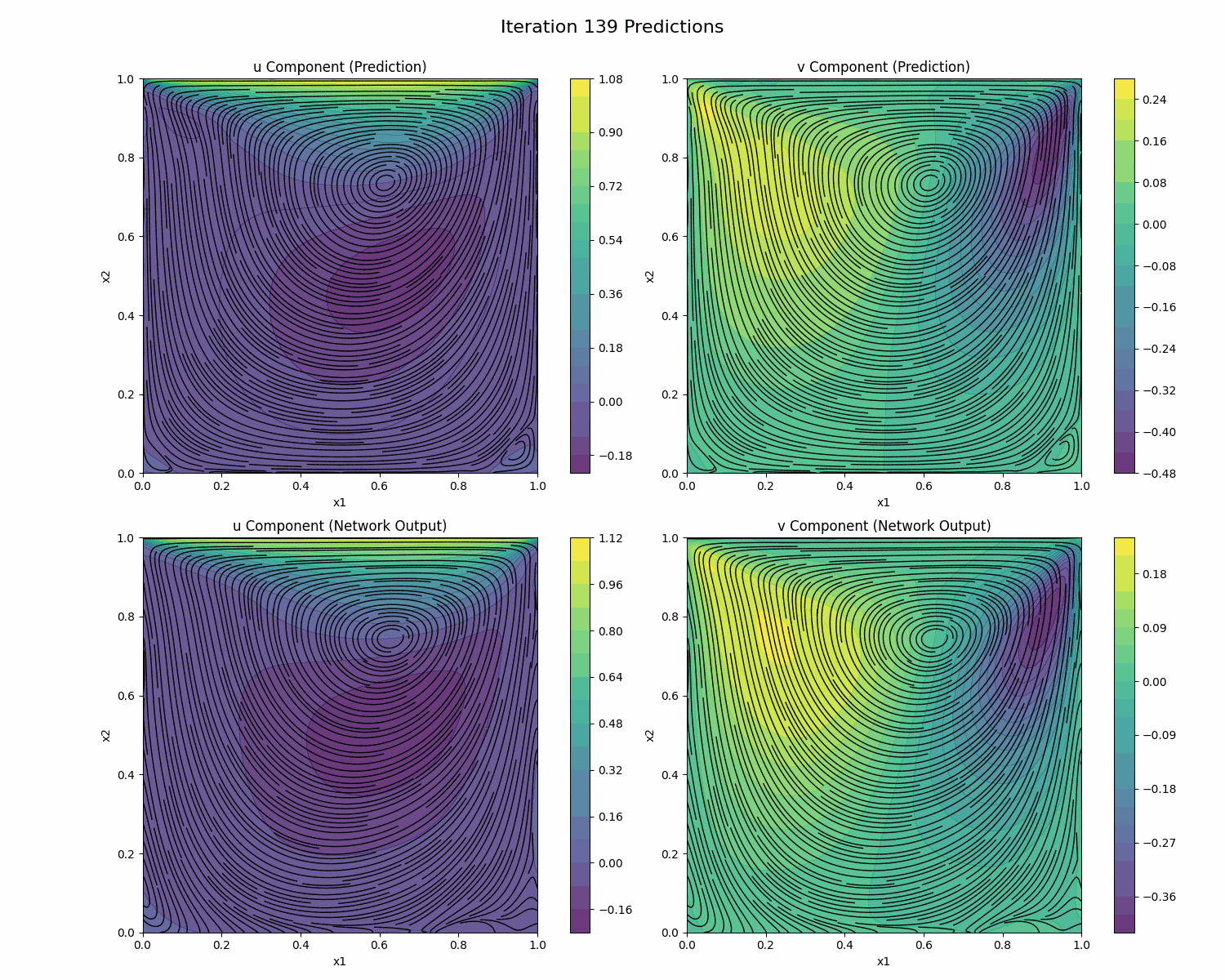}
    \caption{Final converged solution of Re=100 lid-driven cavity flow showing the velocity field at steady state. The bottom shows the result of the first step by PINNs.}
    \label{fig:ns_iteration}
\end{figure}

\begin{figure}[H]
    \centering
    \includegraphics[width=0.8\textwidth]{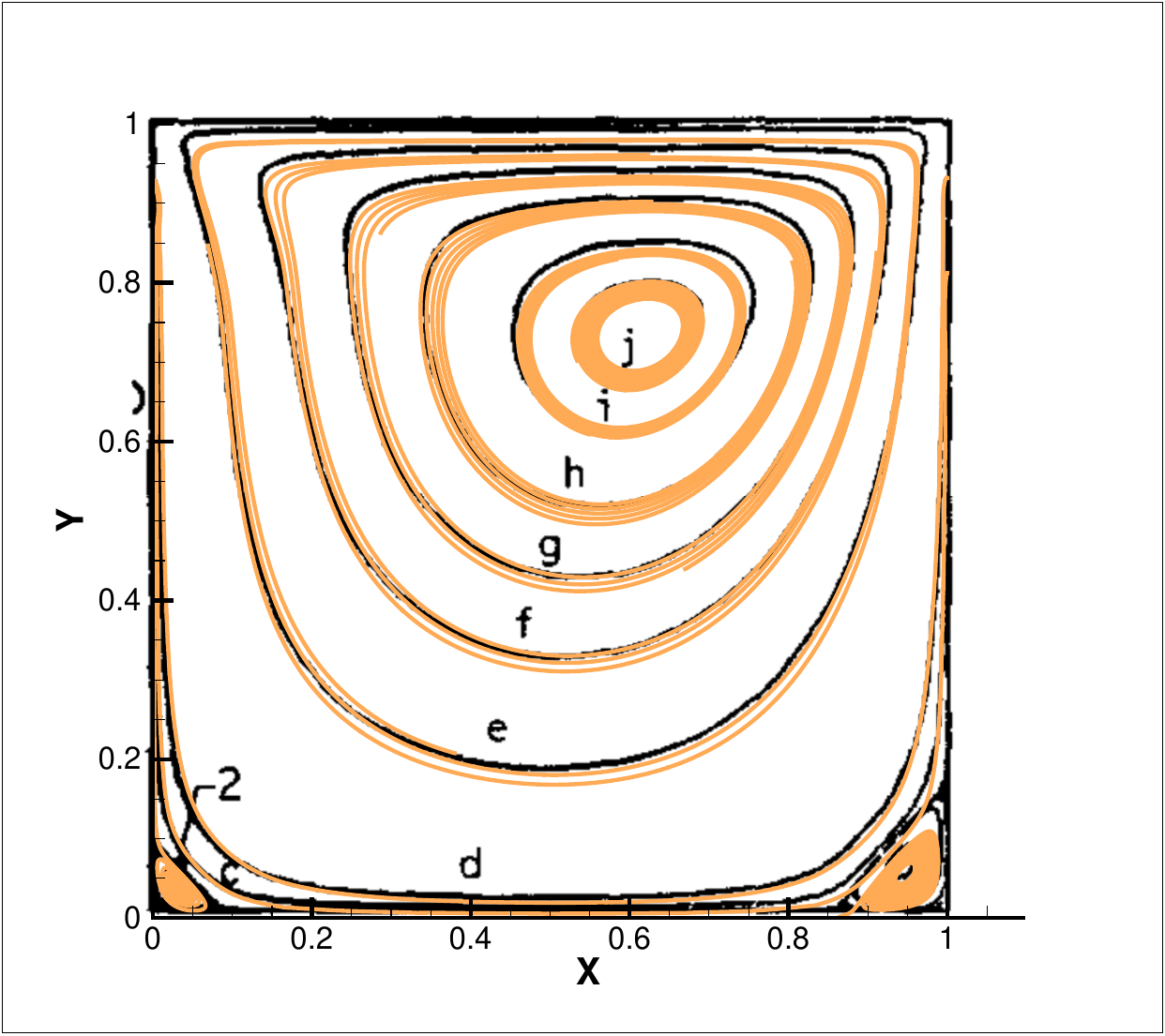}
    \caption{Streamline plot of the Re=100 lid-driven cavity flow, showing the main circulation and corner vortices. Comparison with benchmark results from Ghia et al.~\cite{ghia1982high} demonstrates excellent agreement with significantly fewer computational resources.}
    \label{fig:ns_streamlines}
\end{figure}

Figure \ref{fig:ns_iteration} illustrates the convergence process during the pseudo-time iterations, showing how the flow field evolves toward the steady-state solution. Figure \ref{fig:ns_streamlines} presents the final streamline patterns, which demonstrate excellent agreement with established benchmark results from the literature.

For this example, we used 2880 random sampling points and 6$\times$6 partitions, achieving convergence in 226 seconds. This demonstrates the remarkable efficiency of our approach for complex nonlinear systems like the Navier-Stokes equations, where classical PINNs typically require significantly  much iterations to optimize the solution but still can not clearly capture the details of the flow field, such as the corner vortices and flow lines near the boundaries. The ability to handle the non-trivial pressure-velocity coupling and boundary conditions without explicit mesh generation further highlights the advantages of our hybrid framework.

\section{Theoretical Analysis}
\label{sec:theoretical}

In numerical approximation theory, accuracy and order of convergence constitute fundamental concepts. Accuracy quantifies the approximation error to a target function, typically measured using norms such as $L^2$ or $L^\infty$. Order of convergence characterizes the rate at which this error decreases as the domain is refined through partitioning---a principle known as "divide and conquer." This methodological foundation asserts that complex approximation problems can be systematically decomposed into simpler subproblems. To formalize this concept, we must first establish a precise mathematical definition of function complexity.

\subsection{Function Complexity and Approximation Theory}

Consider a continuous function $f(x)$ defined on an interval $I = [a - \Delta x, a + \Delta x]$. Given a family of polynomial basis functions $\{P_i(x)\}_{i=0}^{\infty}$, we define the complexity of $f$ on $I$ as follows:

\begin{definition}[Function Complexity]
For a given precision threshold $\epsilon > 0$, the complexity $\mathcal{C}_\epsilon(f,I)$ of function $f$ on interval $I$ is defined as the minimum integer $M$ such that there exist coefficients $\{\beta_i\}_{i=0}^M \subset \mathbb{R}$ satisfying:
\[
\left\| f(x) - \sum_{i=0}^M \beta_i P_i(x) \right\|_{L^\infty(I)} \leq \epsilon
\]
\end{definition}

This definition aligns with classical approximation theory and provides a quantitative measure of how "difficult" a function is to approximate using polynomials. A critical observation is that as $\Delta x$ decreases, the complexity $\mathcal{C}_\epsilon(f,I)$ generally decreases as well. In the limit, for any continuous function $f$ and precision $\epsilon$, there exists a sufficiently small interval width such that $f$ can be approximated by a constant function within the desired precision---this is equivalent to the mathematical definition of continuity.

For practical applications, however, achieving high precision (e.g., machine epsilon) using only piecewise constant approximations would require an impractically large number of partitions. This limitation motivates the use of higher-order polynomial bases, which achieve faster convergence rates as the domain is refined. The relationship between partition size, polynomial degree, and approximation error defines the concept of order of accuracy in numerical methods.

\subsection{Polynomial Approximation and Error Sensitivity}

An important consideration when using polynomial approximation is the sensitivity of the approximation error to coefficients. When approximating $f(x)$ as $\sum_{i=0}^M \beta_i P_i(x)$, errors in the coefficients $\beta_i$ propagate to the final approximation with varying sensitivity. Generally, errors in lower-order coefficients (particularly $\beta_1$ for the linear term) have more significant impacts on the overall approximation error.

This sensitivity analysis leads to a critical question regarding deep neural networks (DNNs): Can DNNs accurately approximate even the simplest non-constant function, $y=x$? The answer to this question fundamentally determines whether DNNs can effectively leverage domain partitioning to achieve high-order accuracy. If a DNN can approximate $y=x$ with an error below some threshold $\epsilon_1$, then for any desired accuracy level above $\epsilon_1$, the network could benefit from the "divide and conquer" strategy with efficiency scaling with the number of partitions. Conversely, if DNNs struggle to approximate linear functions with high precision, then domain partitioning would only improve accuracy after the function within each partition becomes nearly constant (i.e., when $\max |f(x) - c_0| \leq \epsilon_1$ for some constant $c_0$).

\subsection{Empirical Analysis of DNNs Approximating Linear Functions}

To investigate this question empirically, we conducted experiments approximating $y=x$ using DNNs with $\tanh$ activation functions, varying both width (1-3 units) and depth (1-3 layers). Theoretically, even a simple network with one hidden unit ($w_1 \tanh(w_2 x)$) can approximate $y=x$ arbitrarily well as $w_1 \to \infty$ and $w_2 \to 0$. However, practical optimization algorithms face significant challenges in finding these parameter values.

Tables \ref{tab:yx_single} and \ref{tab:yx_double} present the approximation errors using ADAM and L-BFGS optimizers in single and double precision, respectively. The results reveal a striking limitation: even for the elementary function $y=x$, DNNs fail to achieve high-precision approximation. Double precision computations improve accuracy by approximately one order of magnitude, while increasing network depth or width yields only marginal improvements.

\begin{table}[H]
    \centering
    \caption{Approximation results for \( y=x \) (Single Precision).}
    \label{tab:yx_single}
    \begin{tabular}{@{}cccc@{}}
        \toprule
        Depth & Width & Adam MAE & L-BFGS MAE \\\\ 
        \midrule
        1 & 1 & 1.64e-02 & 3.44e-04 \\\\ 
        1 & 2 & 1.20e-02 & 1.82e-04 \\\\ 
        1 & 3 & 1.22e-02 & 9.04e-04 \\\\ 
        2 & 1 & 2.03e-02 & 3.60e-04 \\\\ 
        2 & 2 & 2.57e-03 & 4.22e-04 \\\\ 
        2 & 3 & 1.04e-03 & 5.56e-04 \\\\ 
        3 & 1 & 2.39e-02 & 4.37e-04 \\\\ 
        3 & 2 & 1.69e-02 & 3.48e-04 \\\\ 
        3 & 3 & 1.14e-03 & 7.17e-04 \\\\ 
        \bottomrule
    \end{tabular}
\end{table}

\begin{table}[H]
    \centering
    \caption{Approximation results for \( y=x \) (Double Precision).}
    \label{tab:yx_double}
    \begin{tabular}{@{}cccc@{}}
        \toprule
        Depth & Width & Adam MAE & L-BFGS MAE \\\\ 
        \midrule
        1 & 1 & 2.32e-02 & 3.44e-04 \\\\ 
        1 & 2 & 7.16e-03 & 2.76e-04 \\\\ 
        1 & 3 & 8.31e-04 & 4.02e-05 \\\\ 
        2 & 1 & 1.50e-02 & 5.69e-04 \\\\ 
        2 & 2 & 8.46e-03 & 7.30e-05 \\\\ 
        2 & 3 & 2.32e-03 & 7.08e-05 \\\\ 
        3 & 1 & 1.34e-02 & 6.91e-04 \\\\ 
        3 & 2 & 3.11e-03 & 6.99e-04 \\\\ 
        3 & 3 & 2.70e-03 & 3.58e-05 \\\\ 
        \bottomrule
    \end{tabular}
\end{table}

\begin{figure}[H]
    \centering
    \includegraphics[width=0.6\textwidth]{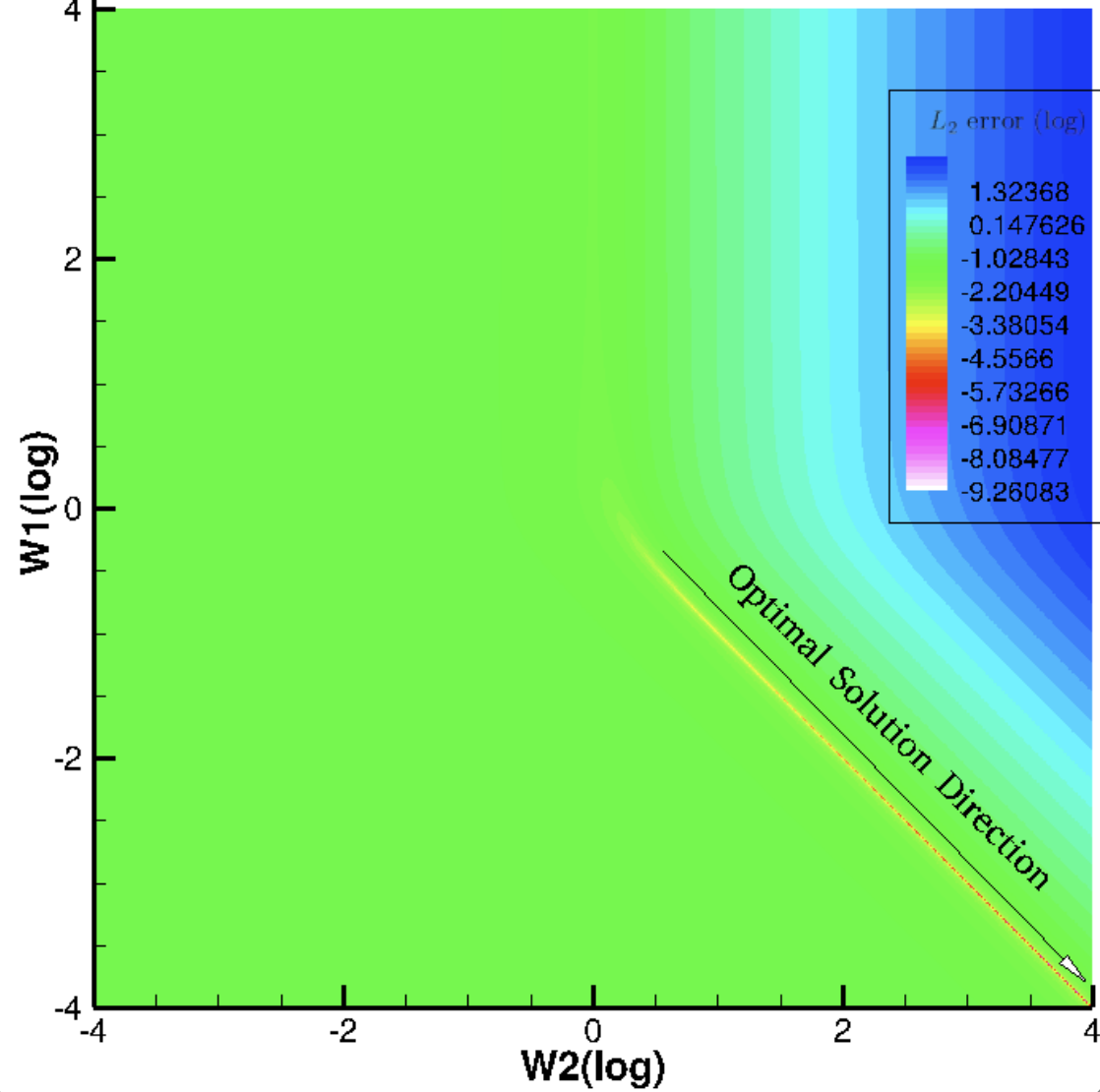}
    \caption{Conceptual optimization landscape for approximating y=x with $w_1 \tanh(w_2 x)$.}
    \label{fig:landscape_linear}
\end{figure}

The fundamental challenge lies in the optimization landscape's geometry. As illustrated in Figure \ref{fig:landscape_linear}, the global minimum for approximating $y=x$ lies along an extremely narrow ridge in the parameter space. Even minor deviations from this ridge can increase the approximation error by orders of magnitude. This inherent difficulty explains why relying solely on DNNs with domain partitioning faces significant limitations in achieving high-order accuracy.

\subsection{Complementary Strengths: DNNs and Polynomial Bases}

Despite their limitations in approximating simple functions with high precision, DNNs excel at maintaining consistent (albeit moderate) accuracy for complex functions with intricate features. Conversely, polynomial approximation achieves high precision for smooth functions but faces challenges with complex local structures, discontinuities, or high-dimensional problems.

Polynomial approximation suffers from several well-documented limitations:
\begin{itemize}
    \item Stability issues such as Runge's phenomenon, where high-order approximations can produce large oscillations
    \item Requirements for high function regularity to achieve theoretical convergence rates
    \item The curse of dimensionality, where the number of basis functions grows exponentially with dimension
\end{itemize}

Our hybrid approach strategically combines the complementary strengths of both methods:
\begin{enumerate}
    \item The DNN component captures complex global features and serves as a stabilizing factor for the approximation
    \item Polynomial basis functions provide the high-order local accuracy and convergence guarantees
    \item Limited domain partitioning further enhances accuracy while maintaining computational efficiency
\end{enumerate}

This synergistic combination transforms the approximation process from relying solely on challenging non-convex optimization to a more tractable two-stage approach: (1) obtaining a reasonable approximation through parameter optimization and (2) refining it through linear space optimization. The theoretical advantages of this hybrid framework are validated by the empirical results presented in Section 3, which demonstrate both high accuracy and computational efficiency across a diverse set of test problems.

\section{Open Source Code}
\includegraphics[width=\textwidth]{picture/logo.png}
All the code related to this paper is open-sourced in the DeePoly repository:

\href{https://github.com/DeePoly/DeePoly}{https://github.com/DeePoly/DeePoly}.

The goal of this project is to develop a general-purpose program for solving PDE equations with high-order accuracy and efficiency. Currently, only the function approximation examples are open-sourced; other examples are being organized. As this project is currently developed and maintained solely by the author Li Liu, \textbf{contributions from interested peers are welcome.}
\section{Conclusion}

This paper proposed a novel framework for scientific computing that combines the strengths of deep neural networks in capturing complex global features with the local convergence accuracy properties of polynomial basis functions. The new method shifts the paradigm from traditional parameter training searching for lower local minima to finding a reasonably good local minimum first, followed by linear optimization within a linear space composed of both DNN features and polynomial basis functions. This approach significantly reduces reliance on non-linear optimization and ensures accuracy through the polynomial basis functions in the linear space, thereby achieving both high efficiency and high accuracy simultaneously.

Despite the promising results demonstrated in our numerical experiments, several challenges remain for future work. 
\begin{enumerate}
    \item First, extensive testing in diverse practical applications is necessary to fully validate the effectiveness of the approach across different domains. 
    \item Second, while our method achieves remarkable accuracy improvements, there are inevitable computational challenges when scaling to high-dimensional problems. Although only a small number of partitions are needed to significantly improve accuracy, the partitioning cost still grows substantially with increasing dimensions. Dynamic partitioning strategies that adaptively refine only in regions requiring higher precision could help alleviate this growth. Additionally, the polynomial basis functions grow exponentially with dimensions, presenting another scalability challenge. Techniques for reducing unnecessary cross-terms in the polynomial expansion could prove valuable for high-dimensional applications. 
    \item Third, while the new method can stably solve discontinuous problems, including convection equations, the accuracy remains insufficient due to the influence of discontinuous regions on the overall matrix properties. By incorporating Spotter's predicted gradients or residuals to adjust equation weights, we expect to significantly improve the solution accuracy in smooth regions of discontinuous problems \cite{pinn_we_reference}.
\end{enumerate}

We believe that, the DeePoly framework represents a significant step toward bridging the gap between traditional numerical methods and deep learning approaches, offering a promising direction for scientific machine learning that combines the best aspects of both paradigms.
\bibliographystyle{plain}
\bibliography{references}

\end{document}